\pdfoutput=1 
\documentclass[11pt]{article}

\usepackage[font=libertinus, citestyle=authoryear]{kurbanlab}

\DeclareAffiliation{hbku}{%
  College of Science and Engineering, Hamad Bin Khalifa University, Doha, Qatar}

\DeclareAffiliation{tamu}{%
  Department of Computer and Electrical Engineering,
  Texas A\&M University, College Station, TX, USA}

\DeclareAffiliation{iub}{%
  Luddy School of Informatics, Computing, and Engineering,
  Indiana University Bloomington, Bloomington, IN, USA}

\usepackage{xspace}
\usepackage{float}
\usepackage{pifont}
\usetikzlibrary{arrows.meta,positioning}

\newsavebox{\fitbox}
\newcommand{\fitwidth}[1]{\sbox{\fitbox}{#1}\ifdim\wd\fitbox>\linewidth\resizebox{\linewidth}{!}{\usebox{\fitbox}}\else\usebox{\fitbox}\fi}

\newcommand{\discuss}[1]{}

\newcommand{\gvb}{GenVidBench\xspace}
\newcommand{\aigvd}{AIGVDBench\xspace}
\newcommand{\genvideo}{GenVideo\xspace}
\newcommand{\dvf}{DVF\xspace}
\newcommand{\rvr}{RvR\xspace}
\newcommand{\rvrstar}{RvR$^{*}$\xspace}
\newcommand{\cmark}{\ding{51}}
\newcommand{\xmark}{\ding{55}}
\DeclareRobustCommand{\hbfull}{\tikz[baseline=-0.6ex]{\fill (0,0) circle (0.75ex);}}
\DeclareRobustCommand{\hbhalf}{\tikz[baseline=-0.6ex]{\fill (0,0) -- (90:0.75ex) arc[start angle=90,end angle=-90,radius=0.75ex] -- cycle; \draw[line width=0.45pt] (0,0) circle (0.75ex);}}
\DeclareRobustCommand{\hbnone}{\tikz[baseline=-0.6ex]{\draw[line width=0.45pt] (0,0) circle (0.75ex);}}
\newcommand{\cistack}[2]{\begin{tabular}[c]{@{}c@{}}#1\\[-1.5pt]{\scriptsize #2}\end{tabular}}
\newcommand{\para}[1]{\noindent\textbf{#1}\hspace{0.5em}}

\crefname{figure}{Figure}{Figures}
\crefname{section}{Section}{Sections}
\crefname{equation}{Equation}{Equations}

\title{Where the Real Clips Came From}
\Subtitle{Corpus Identity in AI-Generated Video Detection}
\RunningTitle{Corpus identity in AI-generated video detection}

\Author{Mert Onur Cakiroglu}{iub}
\Author{Zhihe Lu}{hbku}
\Author{Mehmet Dalkilic}{iub}
\Author[corresponding=hkurban@hbku.edu.qa, orcid=0000-0003-3142-2866]{Hasan Kurban}{hbku}

\CodeURL{https://github.com/KurbanIntelligenceLab/vidaudit}
\ArxivID{2606.31004}

\begin{document}
\makeatletter\pdfstringdef\@pdftitle{Where the Real Clips Came From: Corpus Identity in AI-Generated Video Detection}\makeatother
\maketitle

\begin{abstract}
Benchmarks for detecting AI-generated video take their real clips from a few curated collections, so a clip's label is collinear with where it came from. We call a score's dependence on this origin corpus identity, and we measure it on four benchmarks and six real collections. Three probes read a file's declared frame types, the motion vectors its codec stores, and the frames' appearance. Frame-type counts alone separate real from generated at AUC $0.999$ on \gvb, and re-encoding every clip with one setting at a fixed length still leaves $0.61$ to $0.63$ on three benchmarks. We introduce a fixed frame-type pattern that removes this leak from every clip of at least $2$\,s. The same re-encode lowers the ReStraV detector on \gvb from $0.85$ to $0.60$, just above the $0.59$ its features reach between two real collections, because its $8$ fps repeats frames in real clips. Motion vectors alone separate real collections from one another at AUC $0.51$ to $0.98$, and only one benchmark ships the two labeled real sources needed to measure this separation within it. In deployment, with each threshold set to flag $5\%$ of its training benchmark's real clips, the motion-vector and appearance probes and linear readouts on ReStraV and WaveRep features each flag $15\%$ or more of another real collection. Choosing the threshold on real videos from the target collection brings the rate back to about $5\%$, at the cost of catching fewer generated clips. Clearing the real-vs-real level does not guarantee replication: of the six feature sets whose \gvb score clears it by $0.15$ or more, only WaveRep's two keep their score on \aigvd. A benchmark score is therefore best read together with that level, a score on a second benchmark, and false-positive rates on other real collections. The toolkit and the clip lists of every evaluation set are at \url{https://github.com/KurbanIntelligenceLab/vidaudit}.
\end{abstract}

\section{Introduction}
\label{sec:intro}

AI-generated videos have become realistic enough that people find it hard to tell them apart from real footage~\citep{aigvdbench2024}. Such content now reaches politics~\citep{ramkumar2026}, news~\citep{vaccari2020deepfakes} and finance~\citep{kuznicka2026deepfakes}. Automatic detection systems must therefore keep pace with newly released generators. Progress is evaluated on large benchmarks such as \gvb~\citep{ni2025genvidbenchchallengingbenchmarkdetecting}, with $6.78$M clips, and \aigvd~\citep{aigvdbench2024}, with $440$K clips. However, these benchmarks source their generated clips from a variety of generators but only pull their real clips from one or two sources. Because a clip's label is collinear with its origin, a detector can score well simply by recognizing which collection a clip came from. We call this dependence \emph{corpus identity}. Dataset bias of this kind is an old finding in vision~\citep{torralba2011unbiased}, audits have exposed structural flaws in deepfake datasets~\citep{layton2024sok}, and format and quality shortcuts have been reported for generated images~\citep{grommelt2024fakejpeg,guillaro2025bfree,rajan2025aligned} and video~\citep{fang2026realdatalies,michels2026motionbias}. Yet it remains unclear whether the models are learning to detect generation artifacts or just recognizing the source, how strongly their scores depend on it, which benchmarks allow this to be measured at all, and how this dependence affects a detector in deployment.

\begin{figure}[t]
  \centering
  \fitwidth{%
\definecolor{idc}{RGB}{204,121,167}
\definecolor{genc}{RGB}{0,158,115}
\begin{tikzpicture}[
    font=\sffamily\small,
    >={Stealth[length=2.2mm,width=1.8mm]},
    node/.style={draw=black!70, line width=0.5pt, rounded corners=2pt, minimum height=6.5mm,
                 minimum width=19mm, inner xsep=4pt, fill=white},
    op/.style={font=\sffamily\scriptsize\itshape, text=black!60},
    idpath/.style={->, line width=1.2pt, draw=idc},
    genpath/.style={->, line width=1.2pt, draw=genc},
  ]
  \node[node] (S) at (0, 1.05) {source $S$};
  \node[node] (G) at (0, -1.05) {generated $G$};
  \node[node, draw=idc]  (C) at (3.6, 1.05) {container};
  \node[node, draw=idc]  (M) at (3.6, 0)    {signal};
  \node[node, draw=genc] (A) at (3.6, -1.05) {artifacts};
  \node[node, line width=0.8pt, minimum width=15mm] (D) at (9.0, 0) {score $D$};
  \draw[black!45, line width=0.6pt, dash pattern=on 2pt off 1.5pt] (S) -- node[op, left=2pt] {collinear} (G);
  \draw[idpath]  (S) -- (C);
  \draw[idpath]  (S.east) to[out=-25, in=180] (M.west);
  \draw[genpath] (G) -- (A);
  \draw[idpath]  (C.east) to[out=0, in=150] node[op, above right=0.5pt, pos=0.0] {intervene: re-encode, length filter} (D.north west);
  \draw[idpath]  (M) -- node[op, above, pos=0.45] {measure: floor, or pair} (D);
  \draw[genpath] (A.east) to[out=0, in=-150] (D.south west);
\end{tikzpicture}%
  }
  \caption{\textbf{A benchmark confounds origin with generation.} The label $G$ is collinear with the source $S$, so a score that depends on $S$ through the container or the signal (purple) looks the same as one that detects generation artifacts (green). Italic labels name the operation that controls each channel.}
    \label{fig:model}
\end{figure}

Answering these questions is important because the information about the origin may reach a detector through two routes. The first possible route is the file container, which contains information relating to its encoding, such as its declared frame types and its length. This information can be read before any pixel is decoded (\cref{fig:model}). Bias from such a route can be minimized through a fixed re-encode and a length filter. The second is the decoded signal. This reflects how a collection of clips is filmed, cut, and compressed. Unlike the file container, signal differences lie in the pixels, and removing them would also remove the generation artifacts that detectors should look for. Thus, to verify whether a detector is genuinely recognizing the artifacts it is supposed to, these differences should be measured on such benchmarks. This measurement requires at least two real sources labeled clip by clip. Three of the four public benchmarks we audit lack such a pair (\cref{tab:census}). Among $20$ recent detection papers we surveyed, only one audits the container in full, and none tests in full whether its real sources can be told apart (\cref{app:survey}). In deployment, a detector tuned on one benchmark's real clips can wrongly flag a large share of real clips from another collection (\cref{sec:consequence}).

To audit these channels, we pair three lightweight probes with features from published video detectors and motion representations (\cref{sec:instrument}). Our probes read container frame types, H.264 motion vectors, and CLIP frame appearance~\citep{radford2021clip}. We fit a shared linear classifier called the \emph{readout} on the features of each probe and detector to tell two real sources apart. Its real-vs-real (RvR) area under the receiver operating characteristic curve (AUC) quantifies how much origin information the features expose. We call this measure the \emph{identity floor} when both real collections come from the same benchmark. We call it the \emph{identity distance} when they come from different benchmarks. We evaluate both metrics across six real collections drawn from \gvb, \aigvd, Diffusion Video Forensics (\dvf)~\citep{song2024mmdet}, and \genvideo~\citep{chen2024demamba} augmented with Kinetics-400 (K400)~\citep{kay2017kinetics}.

\begin{figure}[t]
  \centering
  \includegraphics[width=\textwidth]{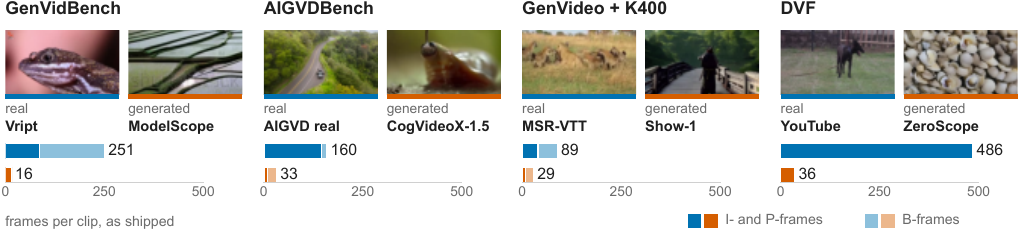}
  \caption{\textbf{Real and generated clips differ before a pixel is decoded.} For each benchmark, a real (blue) and a generated (vermilion) clip near the median length of its class, as shipped. Bar length is the frame count, dark for I-frames (coded alone) and P-frames (predicted from an earlier frame), light for B-frames (predicted from an earlier and a later frame). Frame-type counts alone separate each benchmark's real clips from its generated ones at AUC $0.999$, $0.997$, $0.839$ and $0.893$, left to right (\cref{tab:twobytwo}).}
  \label{fig:teaser}
\end{figure}

\para{Contributions.}
\begin{itemize}[leftmargin=*,nosep]
  \item \textbf{Container leakage and mitigation (\cref{sec:leakage}):} We show that declared frame-type counts alone separate real from generated clips at $0.84$ to $1.00$ AUC across all four benchmarks as shipped (\cref{fig:teaser}). After a standard re-encode and a length filter they still reach $0.61$ to $0.63$ on three of them, because encoders still choose each clip's frame types. We introduce a fixed frame-type pattern for re-encoding that removes this leak from every clip lasting at least $2$\,s. The standard re-encode drops ReStraV's score on \gvb from $0.85$ to $0.60$, just above its identity floor of $0.59$, because the re-encode's $8$ fps gives real clips repeated frames (\citealp{michels2026motionbias}; \cref{sec:floors}).
  \item \textbf{How far apart real collections are (\cref{sec:census}):} Across six real collections, we show that, through codec motion, four collections drawn from YouTube are hard to tell apart (real-vs-real AUC $0.51$ to $0.64$), while \aigvd's real clips and MSR-VTT~\citep{xu2016msrvtt}, which \genvideo distributes at $3$ fps, separate from every other collection at $0.86$ to $0.98$. Adding K400 as a second real pool to \genvideo gives an evaluation set whose codec-motion floor ($0.976$) exceeds the probe's own score ($0.806$), so the set cannot rank detectors.
  \item \textbf{Deployment false-positive shifts and calibration (\cref{sec:consequence}):} We set each decision threshold to a $5\%$ false-positive rate on the training benchmark's real clips. Fitted with our readout, the codec-motion and appearance probes and the features of ReStraV and WaveRep then each flag $15\%$ or more of at least one other real collection, up to $100\%$. Benign changes to a collection's codec or frame rate alone move these rates by tens of percentage points. With every collection processed alike, each rate of $15\%$ or more comes from a detector fitted on \aigvd or falls on \aigvd's real clips. Recalibrating the threshold on real videos of the target collection brings the rate back to about $5\%$.
  \item \textbf{Limits of the identity explanation (\cref{sec:downstream}):} Of the six feature sets whose \gvb score clears their floor by $0.15$ or more, only WaveRep's two keep their score on \aigvd.
  \item \textbf{Toolkit and literature audit (\cref{sec:toolkit,sec:conclusion}):} We release the VidAudit toolkit, the clip lists of every evaluation set, actionable reporting recommendations for benchmarks and detectors, and a survey of $20$ recent detection papers for six of the controls we use (\cref{app:survey}).
\end{itemize}

\section{Related Work}
\label{sec:related}

\para{Dataset bias and shortcut learning.} Shortcuts created by how a dataset is built are known in vision, generated images, and audio-visual deepfakes, and recent work finds them in generated video. \citet{torralba2011unbiased} showed that vision datasets are identifiable from their images, and \citet{liu2025decade} found that modern networks still tell large image datasets apart with high accuracy. \citet{grommelt2024fakejpeg} found that detectors trained on a widely used generated-image benchmark learn JPEG compression and image size, which differ between that benchmark's real and generated images. This is the image counterpart of our container channel (\cref{sec:chain}). Video forensics has long read the container's group-of-pictures structure to detect double compression~\citep{wang2006double,vazquezpadin2012gop}. B-Free~\citep{guillaro2025bfree} and aligned datasets~\citep{rajan2025aligned} pair each real image with its own reconstruction, so that content and format cannot separate the classes. \citet{smeu2025shortcuts} found a leading-silence shortcut in audio-visual deepfake datasets. For generated video, Real Data Lies~\citep{fang2026realdatalies} names a quality shortcut, and G2VD~\citep{du2026g2vd} trains against generator bias with reconstructed counterfactuals. \citet{michels2026motionbias} show that three of four motion-based detectors exploit preprocessing and motion-magnitude bias and fall to chance on a rebalanced set. They also show that ReStraV's frame sampler repeats frames in clips below $12$ fps or shorter than its $2$\,s window, and that a single feature which detects these repeats comes close to ReStraV's accuracy. These shortcuts all make a score depend on where its clips came from. We measure that dependence across benchmarks, show where it cannot be measured, and follow it into deployment.

\para{Evaluation audits.} Re-evaluations under a fair protocol have shown standard scores to overstate progress. \citet{choe2020evaluating} found years of progress in weakly supervised localization vanish, and \citet{cheng2026sanity} showed that cross-domain AUC hides mixed-domain failure in face forgery detection. We audit what the scores of a younger field measure.

\para{Detecting AI-generated video.} Video detectors differ in what they read. Some read per-frame features and their temporal structure (D3~\citep{ma2024d3}, DeCoF~\citep{maaz2024decof}), some compare visual and text trajectories (ATSS~\citep{xu2024atss}, CMTA~\citep{zhao2024cmta}), and some fine-tune a multimodal language model (VidGuard-R1~\citep{park2025vidguard}). ReStraV~\citep{restrav2024} reads the straightness of a deep-feature trajectory, and WaveRep~\citep{corvi2025waverep} reads pixel-level forensic traces. Motion features can also come from outside detection, from the descriptor of the FVMD quality metric~\citep{liu2024fvmd} or from optical flow~\citep{teed2020raft}. We re-score D3, ReStraV, WaveRep, and the two motion representations under one readout, survey the other detectors (\cref{app:survey}), and audit the benchmarks themselves.

\section{Measuring Corpus Identity}
\label{sec:instrument}

\subsection{A Benchmark Confounds Origin with Generation}
\label{sec:chain}

In a detection benchmark a clip's label and its origin are collinear, so a score can depend on origin through channels that bypass generation. Let $G$ say whether a clip was generated and $S$ which source it was drawn from, either a real collection such as Vript~\citep{yang2024vript} or a generator such as Pika. A benchmark labels clips by $G$, so it cannot tell a score $D$ that depends on $S$ from one that detects generation. We call the dependence of $D$ on $S$ \emph{corpus identity}. It is the dataset bias of \citet{torralba2011unbiased} in a two-class setting, where one class is a set of real collections and the other a set of generators.

\cref{fig:model} draws the two channels through which identity can reach a score, the \emph{container} and the \emph{signal} of \cref{sec:intro}. A channel that carries identity can be intervened on, measured, or paired. The container can be intervened on. Its encoder settings and length are properties of the file, and a re-encode and a length filter can set them to one value for every clip. The signal cannot be set this way, because setting it to one value would remove the artifacts a detector reads. It can only be measured, by contrasting real clips of different origin, or paired, by building the generated side from the same real clips.

\subsection{Video Coding in Brief}
\label{sec:coding}

The container and codec-motion probes rest on how H.264 stores a clip, as three types of frame linked by motion vectors. An I-frame is coded on its own. A P-frame is predicted from an earlier frame, and a B-frame from an earlier and a later one. Both store mainly what changed, and together we call them the non-I-frames. To predict a frame, the encoder splits it into $16{\times}16$-pixel blocks called macroblocks. For each macroblock it records a motion vector, the offset of the best-matching block in the reference frame. The run of frames from one I-frame to the next is a group of pictures (GOP). Each frame's type is declared in its header, so the counts of each type belong to the container and can be read without reconstructing a pixel. The libx264 encoder~\citep{x264} chooses frame types itself unless told otherwise. Its constant rate factor (CRF) sets a quality target, which the encoder meets with fewer bits on simpler content. Some clips ship in MPEG-4 Part 2, an older codec with the same frame types.

\subsection{Cells, Readout, and Pipelines}
\label{sec:cells}

A \emph{cell} is a fixed set of real and generated clips on which scores are computed. The real clips that a cell draws from one collection form a \emph{real pool}. The \gvb cell takes $3{,}000$ clips from each of seven of \gvb's eight generators and from each of its two real pools, Vript and HD-VG-130M~\citep{wang2023videofactory}, for $27{,}000$ clips in all. \aigvd ships seven generators. Its cell holds $1{,}396$ real clips and $888$ generated clips from the five generators whose clips are long enough for the length filter defined below. The \emph{third cell} combines \genvideo's generators with MSR-VTT and K400 real pools (\cref{sec:census}). The \emph{census} is a separate sample of $500$ clips per source from all four benchmarks, on which we run the container probe and compare real collections (\cref{sec:leakage,sec:census}). A fourth cell holds \dvf's census clips.

Every score comes from one shared \emph{readout}, which we re-fit for each feature set. The readout applies median imputation and standardization, then an $L_2$-regularized logistic regression with balanced classes. Scores are means over five seeds unless a table marks them as a single seed, and intervals come from bootstrap resampling (\cref{app:ci}). Scores are AUCs under leave-one-generator-out (LOGO) evaluation, in which each generator is held out in turn. LOGO out-of-distribution (OOD) trains on real clips and the other generators, then scores the held-out generator against held-out real clips. We use a detector's name for its feature set under this readout.

The \emph{deep detectors} are the feature sets that a deep network reads from decoded frames. They are D3, ReStraV, two feature sets from WaveRep, and CLIP, which also serves as our appearance probe. WaveRep's released classifier scores each frame with a log-likelihood ratio. \emph{WaveRep's head} is the readout on the mean, median, and maximum of these scores over the clip. \emph{WaveRep's backbone} is the readout on the classifier's $768$-dimensional frame features, averaged over the clip.

A clip is read through one of three \emph{pipelines}. \emph{As shipped} is the benchmark's own file. The \emph{recipe} re-encodes it with libx264 at CRF $23$, with a GOP of $12$ frames, at $8$ fps, and at $512{\times}512$ pixels. The \emph{length filter}, applied after the recipe, keeps only clips with at least $K{=}12$ non-I-frames that carry motion vectors and reads the first $K$ of them (\cref{app:cells}). The \emph{native pipeline} is a $512{\times}512$ H.264 re-encode of the \gvb cell at each clip's own frame rate. The container and codec-motion probes read the recipe on every cell, with one exception: in \cref{tab:identity} the codec-motion probe reads the \gvb cell through the native pipeline. We score the deep detectors on clips as shipped, as the D3 and ReStraV papers do, and again with every clip through the recipe (\cref{sec:floors,sec:reprocess}). The recipe drops the two shortest \gvb generators, Stable Video Diffusion and MuseV, so five remain.

\begin{table}[t]
  \centering
  \footnotesize
  \setlength{\tabcolsep}{4pt}
  \renewcommand{\arraystretch}{1.05}
  \caption{\textbf{Only \gvb ships two labeled real sources.} Native codec, frame rate (real / generated) and median length, from $500$ clips per source. Floor: codec-motion \rvr between the two real pools under the recipe; for a corpus with one real pool, the \rvrstar to the other corpora. Gen.: generators with at least $100$ clips after the length filter. Under the native pipeline of \cref{tab:identity} the \gvb floor is $0.643$. R1 and Adm.\ (admitted by the gates) refer to the cell.}
    \label{tab:census}
  \fitwidth{%
  \begin{tabular}{@{}llrlrccc@{}}
    \toprule
    Corpus & Labeled real sources & Gen. & Codec / fps (real / gen.) & Length & Floor or \rvrstar & R1 & Adm. \\
    \midrule
    \gvb 27k~\citep{ni2025genvidbenchchallengingbenchmarkdetecting} & Vript, HD-VG-130M & 7 (5 at $8$ fps) & H.264 30 / H.264 10 & $250$ / $16$ & $0.54$ [$0.50,0.66$] & \cmark & \cmark \\
    \aigvd~\citep{aigvdbench2024} & one (OpenVid-HD) & 5 (7 shipped) & H.264 30 / H.264 24 & $158$ / $33$ & \rvrstar $0.86$--$0.97$ & \xmark & \xmark \\
    \genvideo + K400~\citep{chen2024demamba} & MSR-VTT (K400 ours) & 8 (10 shipped) & H.264 3--30 / H.264 8 & $89$ / $29$ & $0.976$ & \cmark & \xmark \\
    \dvf~\citep{song2024mmdet} & one (YouTube) & 8 (6 paired) & MPEG-4 30 / H.264 8 & $484$ / $36$ & \rvrstar $0.51$--$0.96$ & \xmark & \xmark \\
    \bottomrule
  \end{tabular}}
\end{table}

\subsection{Three Probes}
\label{sec:probes}

Each probe reads one channel of \cref{fig:model} and is fitted with the shared readout.

\para{Container probe.} The container probe counts the I-, P-, and B-frames that a clip's frame headers declare. After the length filter, it counts only the frames up to the $K$th non-I-frame that carries motion vectors, which are the frames the codec-motion probe reads. A one-feature variant reads the total frame count alone. If the probe scores above chance, the container still carries identity.

\para{Codec-motion probe.} The codec-motion probe reads thirteen statistics of the stored motion vectors over the same $K$ frames. Four measure how much motion there is: the mean magnitude, the share of zero vectors, the variance, and the mean change from frame to frame. The other nine measure how the motion is structured over time. They summarize the spectrum, autocorrelation, and acceleration of the motion in each region of a $4{\times}4$ grid (\cref{app:probes}).

\para{Appearance probe.} The appearance probe averages CLIP ViT-B/32~\citep{radford2021clip} features of the decoded frames over the clip and reduces them by principal component analysis (PCA) to $13$ dimensions, the dimension of the codec-motion probe.

\subsection{Floor, Margin, and Distance}
\label{sec:floor}

For a cell with two real pools, the \emph{identity floor} of a feature set is its real-vs-real (\rvr) AUC. It is the AUC of the readout trained to tell the two real pools apart, scored on held-out real clips. The two pools share $G$ and differ only in $S$, so whatever the readout finds between them is origin. The floor therefore shows how much origin the features expose, for the collections the two pools cover. It cannot show whether a detector trained on those features relies on that origin (\cref{sec:originfpr}). It is a reference level for the detection score and places no bound on it. A cell with one real pool has no floor. The floor must be measured on the same features and readout as the score. On the \gvb cell, CLIP's raw $512$-dimensional features give a floor of $0.845$ and its $13$-dimensional reduction $0.766$.

A feature set's \emph{margin} is its LOGO OOD minus its floor. The two AUCs come from different tasks, real versus generated and real versus real, so the margin serves only as a screen. A score passes this screen only if it clears its floor by $0.15$ or more, and moving this cut anywhere between $0.10$ and $0.20$ changes the result for only one feature set in \cref{tab:identity}, the codec-motion probe (\cref{app:gates}). Even a score that clears the cut can still carry identity from collections that the two pools do not cover, or from processing they share, such as the frame rate and codec common to \gvb's two pools.

Computed between real pools from two different corpora, at a common sample size, the \rvr AUC becomes the \emph{identity distance} \rvrstar between the two collections. It measures how far apart two collections are, and unlike the floor it defines no margin.

\subsection{Gates and Controls}
\label{sec:gates}
\label{sec:protocol}

Three \emph{gates}, fixed before any detector ran on a new cell, decide whether the cell can rank detectors. Gate R1 asks for two independently curated real sources with per-clip labels, so that a floor exists (\cref{tab:census}). Gate R2 asks for at least five generators with at least $100$ clips each after the length filter. Gate R3 asks for a codec-motion floor below $0.90$, because two real pools that the probe separates almost perfectly make the floor uninformative. A cell that fails a gate is reported and receives no ranking. Like the margin cut, the gate thresholds are diagnostic screens fixed in advance, and no significance level attaches to them. \cref{tab:controls} lists every control, the check fixed in advance for it, and what a corpus must supply.

\section{The Container Channel}
\label{sec:leakage}

\begin{table}[t]
  \centering
  \footnotesize
  \setlength{\tabcolsep}{3.5pt}
  \renewcommand{\arraystretch}{1.05}
  \caption{\textbf{Only a fixed frame-type pattern with a tail buffer brings the container probe to chance everywhere.} LOGO OOD AUC of the I-, P-, and B-frame counts on the census ($500$ clips per source). Columns: as shipped, with the length filter, with the recipe's re-encode, with both, and with both plus a fixed frame-type pattern and a three-frame tail buffer. Brackets: $95\%$ intervals. Bold: the interval contains chance. With the fixed pattern and the tail buffer every clip has the same counts, so $0.500$ holds by construction.}
    \label{tab:twobytwo}
  \fitwidth{%
  \begin{tabular}{@{}lccccc@{}}
    \toprule
    Corpus & Shipped & Length filter & Re-encode & Both & Fixed pattern, tail buffer \\
    \midrule
    \gvb & $0.999$ & $\mathbf{0.578}$ [$0.37,0.71$] & $0.975$ & $\mathbf{0.481}$ [$0.41,0.55$] & $\mathbf{0.500}$ \\
    \aigvd & $0.997$ & $0.945$ & $0.929$ & $\mathbf{0.633}$ [$0.48,0.79$] & $\mathbf{0.500}$ \\
    \genvideo + K400 & $0.839$ & $0.704$ & $0.952$ & $0.613$ [$0.57,0.67$] & $\mathbf{0.500}$ \\
    \dvf & $0.893$ & $0.785$ & $0.800$ & $0.625$ [$0.54,0.70$] & $\mathbf{0.500}$ \\
    \bottomrule
  \end{tabular}}
\end{table}

As shipped, frame-type counts separate real from generated on every corpus, and no single fix brings the probe to chance on all four (\cref{tab:twobytwo}). The probe reaches $0.84$ to $1.00$ for two reasons. Every corpus combines long real clips with short generated ones and encodes the two sides at different frame rates. Each fix on its own leaves one of these open. On \aigvd the length filter alone removes the difference in length, and the probe on the total frame count falls to AUC $0.496$. The probe on all three counts still reaches $0.945$, because among the twelve non-I-frames that the filter keeps, real clips carry $8.7$ P-frames to $3.3$ B-frames and generated clips the reverse. The re-encode alone makes the frame structure nearly uniform but leaves the difference in length, and the probe reaches $0.929$. Together the two fixes bring \gvb to chance ($0.481$) and leave $0.61$ to $0.63$ on the other three corpora, where only \aigvd's interval reaches chance.

What survives both fixes is the encoder's own choice of frame types, which a fixed frame-type pattern and a tail buffer together remove. After both fixes, real \dvf clips still carry about one more P-frame among their twelve non-I-frames than generated clips ($4.4$ against $3.4$). The reason is that libx264 decides for each clip, from its content, where to place B-frames and where a shot change calls for a fresh I-frame. A fixed frame-type pattern takes this choice away from the encoder. It allows no shot-change I-frames, places two B-frames after every I- or P-frame, and uses a GOP of $12$. This pattern brings the probe to chance on \gvb and \dvf, to $0.70$ on \aigvd, and to $0.55$ on \genvideo (\cref{tab:supp_container_grid}). The rest comes from the end of the stream, where libx264 rewrites the final B-frames as P-frames. For clips that barely pass the filter, the twelve non-I-frames the probe reads therefore end differently depending on the clip's length. Keeping only clips with a \emph{tail buffer} of three frames past them, which means clips of at least $2$\,s, brings the probe to exactly $0.500$ on all four corpora. The buffer is itself a length filter. It removes every clip of AnimateDiff and Open-Sora on \aigvd and of VideoCrafter2 on \gvb, and it keeps $2{,}033$ of $3{,}033$ clips on \aigvd (\cref{app:container}).

Outside this section the codec-motion probe never reads the fixed pattern, so its frame types still vary from clip to clip. Its floors and distances in \cref{sec:census} therefore include whatever identity the frame types carry.

\section{The Signal Channel}
\label{sec:census}

\begin{figure}[t]
  \centering
  \includegraphics[width=0.80\textwidth]{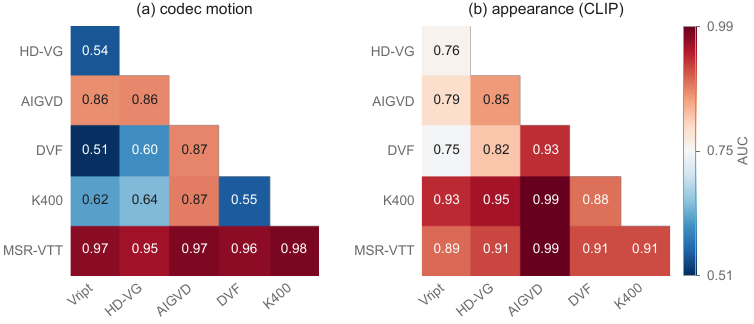}
  \caption{\textbf{Real collections separate from one another.} Real-vs-real AUC between six real pools after the recipe, through (a) the thirteen codec-motion features and (b) CLIP, with $463$ to $500$ clips per pool. Red marks pairs that are easy to tell apart. Bootstrap $95\%$ intervals are at most $\pm0.09$ for codec motion, and the seed spread is at most $0.04$ for CLIP. HD-VG: HD-VG-130M; K400: Kinetics-400; \aigvd and \dvf: those benchmarks' real pools.}
  \label{fig:census}
\end{figure}

\para{Census of real collections.} Codec motion alone tells real collections apart, most strongly those collected or processed differently from the rest. The census holds six real pools: Vript and HD-VG-130M from \gvb, \aigvd's pool (drawn from OpenVid-HD~\citep{aigvdbench2024}), MSR-VTT~\citep{xu2016msrvtt} and Kinetics-400 from the third cell, and the YouTube pool of \dvf. Each was put through the recipe. \cref{fig:census} gives the \rvr AUC for every pair of pools, which is a floor within a corpus and an identity distance across corpora. The four YouTube pools distributed at $30$ fps (Vript, HD-VG-130M, Kinetics-400, and \dvf's pool) sit at $0.51$ to $0.64$ from one another. \aigvd's pool and MSR-VTT, by contrast, are told apart from every other pool at $0.86$ to $0.98$. MSR-VTT also comes from YouTube, but \genvideo distributes it at $3$ fps and $298{\times}224$ pixels (\cref{app:censusdetail}).

Appearance separates the four YouTube pools more than codec motion does, at $0.75$ to $0.95$ through CLIP, and any two of the six pools at $0.75$ to $0.99$.

\begin{table}[t]
  \centering
  \footnotesize
  \setlength{\tabcolsep}{5pt}
  \renewcommand{\arraystretch}{1.02}
  \caption{\textbf{Feature sets with similar scores expose different amounts of origin.} Floor: \rvr between \gvb's two real pools through each feature set. Margin: LOGO OOD minus floor; bold floors are cleared by less than $0.15$. Right: LOGO OOD re-fitted on \aigvd, with $95\%$ intervals from resampling its held-out generators at a single seed, and $\Delta$, the \aigvd score minus the \gvb score (\cref{sec:transfer}). Upper rows: a single seed (five-seed \gvb means in \cref{app:ci}), deep detectors on clips as shipped and the codec-motion probe on the native pipeline. $\dagger$: \aigvd read through the recipe; with \gvb also through the recipe the probe falls from $0.736$ to $0.602$. WaveRep's backbone on \aigvd is the five-seed mean of \cref{tab:recipe_paired}, without an interval. Lower block: every clip through the recipe, five-seed means.}
    \label{tab:identity}
  \fitwidth{%
  \begin{tabular}{@{}lrccccc@{}}
    \toprule
    Detector (feature source) & Dim. & \gvb LOGO OOD & Floor (\rvr) & Margin & Re-fitted on \aigvd [$95\%$ CI] & $\Delta$ \\
    \midrule
    D3 features~\citep{ma2024d3} & 768 & $0.557$ & $\mathbf{0.546}$ & $+0.011$ & $0.722$ [$0.686,0.754$] & $+0.165$ \\
    CLIP ViT-B/32, PCA-13 (appearance probe) & 13 & $0.852$ & $\mathbf{0.766}$ & $+0.086$ & $0.750$ [$0.701,0.828$] & $-0.102$ \\
    WaveRep head~\citep{corvi2025waverep} & 3 & $0.996$ & $0.534$ & $+0.462$ & $0.981$ [$0.959,0.998$] & $-0.015$ \\
    RAFT flow, 13 statistics~\citep{teed2020raft} & 13 & $0.855$ & $0.627$ & $+0.228$ & $0.648$ [$0.622,0.679$] & $-0.207$ \\
    Codec-motion probe, 13-d & 13 & $0.832$ & $0.643$ & $+0.189$ & $0.602$ [$0.528,0.676$] & $-0.230^{\dagger}$ \\
    ReStraV~\citep{restrav2024} & 21 & $0.931$ & $0.586$ & $+0.345$ & $0.673$ [$0.422,0.894$] & $-0.258$ \\
    FVMD~\citep{liu2024fvmd} & 1024 & $0.880$ & $0.574$ & $+0.306$ & $0.606$ [$0.566,0.659$] & $-0.274$ \\
    \midrule
    WaveRep backbone & 768 & $0.999$ & $0.777$ & $+0.222$ & $0.999$ & $+0.000$ \\
    \midrule
    \multicolumn{7}{@{}l}{\emph{Every clip through the recipe, five \gvb generators}} \\
    CLIP ViT-B/32, PCA-13 & 13 & $0.919$ & $0.752$ & $+0.168$ & $0.629$ [$0.592,0.643$] & $-0.290$ \\
    WaveRep head & 3 & $0.971$ & $0.539$ & $+0.432$ & $0.970$ [$0.941,0.987$] & $-0.001$ \\
    WaveRep backbone & 768 & $0.984$ & $0.778$ & $+0.206$ & $0.987$ [$0.977,0.994$] & $+0.003$ \\
    ReStraV & 21 & $0.601$ & $\mathbf{0.587}$ & $+0.014$ & $0.783$ [$0.655,0.937$] & $+0.182$ \\
    \bottomrule
  \end{tabular}}
\end{table}

\para{One corpus in four can measure its floor.} Only \gvb ships the two independently curated real sources with per-clip labels that a floor needs (gate R1, \cref{tab:census}). \aigvd ships one real source. \genvideo's validation split ships only MSR-VTT, so Kinetics-400, the third cell's second pool, is our addition. \dvf names two sources, but every real clip in its release carries a YouTube id, so the two cannot be told apart clip by clip (\cref{app:censusdetail}). \gvb's two pools also come from YouTube, but they are separate releases with per-clip labels. On three of the four corpora as shipped, the floor is therefore undefined, and a detector's margin cannot be reported.

\para{The third cell fails its gate.} We built the third cell to supply a second floor, from \genvideo's eight generators and the MSR-VTT and Kinetics-400 pools ($4{,}906$ clips after the filter), under a construction protocol fixed before any detector ran (\cref{app:cells}). It passes R1 and R2 but fails R3, because its codec-motion floor of $0.976$ exceeds the gate's $0.90$. The floor even exceeds the probe's own LOGO OOD of $0.806\pm0.011$ over five seeds. The cell therefore receives no ranking, a verdict that describes the cell and claims nothing about any detector.

\para{Floors through each detector's features.}
\label{sec:floors}
Feature sets with similar scores sit at very different floors on \gvb (\cref{tab:identity}). The appearance probe and D3's features clear their floors by less than $0.15$. The motion and trajectory feature sets (RAFT, the codec-motion probe, FVMD, and ReStraV) clear them by $0.19$ to $0.35$. WaveRep's head reaches $0.996$ at a floor of $0.534$, while WaveRep's backbone reaches $0.999$ at a floor of $0.777$. Reading one network's final scores or its internal features can thus change how much identity it exposes, although the backbone's higher dimension may contribute, as dimension does for CLIP (\cref{sec:floor}). We report the table with no ranking intended.

Putting every clip through the recipe removes ReStraV's margin and leaves WaveRep's (lower block of \cref{tab:identity}). Restricted to the clips and five generators that survive the recipe (\cref{tab:recipe_paired}), ReStraV falls from $0.852$ as shipped to $0.601$ under the recipe, just above its floor of $0.587$. Frame rate is the processing step that moves ReStraV (\cref{sec:reprocess}), through the repeated frames that \citet{michels2026motionbias} found (\cref{sec:related}). As shipped, four of the five \gvb generators that survive the recipe run at $8$ to $10$ fps and the real clips at $30$ fps, so repeats occur only on the generated side (\cref{app:censusdetail}). Under the recipe's $8$ fps every clip has them. RAFT and FVMD also read each class of \gvb at its own frame rate, and we did not re-score them under the recipe. WaveRep's head moves from $0.995$ to $0.971$ and keeps a margin of $0.43$.

\para{Matching and pairing.} Two adversaries match the generated side to the real side within each benchmark (\cref{app:adaptive}). Matching on motion magnitude moves no feature set by more than $0.016$. Matching on the codec-motion probe's own score lowers FVMD by $0.10$ and RAFT by $0.06$ and moves the others by at most $0.02$. Scoring each fake against its own source clip lowers no AUC by more than $0.031$.

\section{Identity at Deployment}
\label{sec:consequence}

\subsection{False Positives Depend on Where a Real Video Came From}
\label{sec:originfpr}

\begin{table}
  \centering
  \footnotesize
  \setlength{\tabcolsep}{3pt}
  \renewcommand{\arraystretch}{1.02}
  \caption{\textbf{Every feature set over-flags some real collection.} Each feature set is thresholded at $5\%$ false positives on its training cell's held-out real clips and applied to up to $500$ real clips per collection, as shipped for the deep detectors and under the recipe for the codec-motion probe. Rates are means over five seeds. Across seeds a deep-detector rate has a standard deviation of at most $0.06$, and a codec-motion rate spans at most $0.16$. On $500$ clips a rate of $0.05$ has a binomial $95\%$ interval of $[0.03,0.07]$. Gray: the training cell's own pools. Bold: three times nominal or more. HD-VG: HD-VG-130M; K400: Kinetics-400.}
    \label{tab:originfpr}
  \fitwidth{%
  \begin{tabular}{@{}llcccccc@{}}
    \toprule
    Detector & Fitted on & Vript & HD-VG & \aigvd & MSR-VTT & K400 & \dvf \\
    \midrule
    Codec-motion & \aigvd & $\mathbf{.28}$ & $\mathbf{.28}$ & \textcolor{gray}{$.05$} & $\mathbf{.26}$ & $\mathbf{.31}$ & $\mathbf{.30}$ \\
    Codec-motion & \dvf & $.05$ & $.04$ & $\mathbf{.22}$ & $.04$ & $.06$ & \textcolor{gray}{$.05$} \\
    Codec-motion & \genvideo & $.10$ & $.07$ & $\mathbf{.16}$ & \textcolor{gray}{$.00$} & \textcolor{gray}{$.09$} & $.10$ \\
    Codec-motion & \gvb & \textcolor{gray}{$.05$} & \textcolor{gray}{$.05$} & $\mathbf{.17}$ & $.04$ & $.07$ & $.07$ \\
    \midrule
    CLIP & \gvb & \textcolor{gray}{$.04$} & \textcolor{gray}{$.07$} & $\mathbf{.17}$ & $.04$ & $.01$ & $.05$ \\
    CLIP & \aigvd & $.05$ & $.03$ & \textcolor{gray}{$.06$} & $.07$ & $.04$ & $.05$ \\
    WaveRep head & \gvb & \textcolor{gray}{$.05$} & \textcolor{gray}{$.04$} & $.07$ & $.07$ & $.03$ & $\mathbf{.27}$ \\
    WaveRep head & \aigvd & $.04$ & $.03$ & \textcolor{gray}{$.06$} & $.05$ & $.02$ & $\mathbf{.24}$ \\
    WaveRep backbone & \gvb & \textcolor{gray}{$.04$} & \textcolor{gray}{$.05$} & $.04$ & $\mathbf{.81}$ & $\mathbf{.15}$ & $\mathbf{.33}$ \\
    WaveRep backbone & \aigvd & $.10$ & $.12$ & \textcolor{gray}{$.05$} & $\mathbf{1.00}$ & $\mathbf{.60}$ & $\mathbf{.48}$ \\
    ReStraV & \gvb & \textcolor{gray}{$.04$} & \textcolor{gray}{$.05$} & $.05$ & $\mathbf{.28}$ & $.03$ & $.05$ \\
    ReStraV & \aigvd & $.09$ & $.12$ & \textcolor{gray}{$.05$} & $\mathbf{.90}$ & $.10$ & $.12$ \\
    \bottomrule
  \end{tabular}}
\end{table}

Thresholded at $5\%$ false positives on its own benchmark, every feature set in \cref{tab:originfpr} flags some public real collection at three times that rate or more. We fit each detector on one cell, the deep detectors on clips as shipped and the codec-motion probe under the recipe. We set its threshold on that cell's held-out real clips and apply it to up to $500$ real clips from each of six public collections. On \gvb's own two pools, by contrast, each feature set's rates differ by at most $0.03$, although the floors of CLIP and WaveRep's backbone reach $0.77$ and $0.78$ (\cref{tab:identity}).

The largest rates fall on collections that are far from the training pool or were processed differently. MSR-VTT, which \genvideo distributes at $3$ fps although the original clips run at $25$ to $30$ fps~\citep{michels2026motionbias}, draws the highest. WaveRep's backbone flags $81\%$ of it when fitted on \gvb and $100\%$ when fitted on \aigvd, and ReStraV fitted on \aigvd flags $90\%$. Frame rate alone moves ReStraV's rate, because at $3$ fps most of the $24$ frames it reads are repeats (\cref{sec:reprocess,app:harness}). MSR-VTT's low frame rate and small frames also resemble the generated side more than any other real pool does. A detector that learned how generated clips are processed would therefore flag MSR-VTT too, and this table cannot tell that detector from one that learned its training cell's real pool. The codec-motion probe fitted on \aigvd, whose real pool is far from every other collection (\cref{sec:census}), flags $26$ to $31\%$ of each of them.

\subsection{Re-processing Moves the Rate}
\label{sec:reprocess}

Which processing step moves a false-positive rate depends on what the detector reads (\cref{tab:reprocess}). We keep each threshold of \cref{tab:originfpr} and score the same real clips after three steps, each adding one change to the last: a plain H.264 re-encode at the native frame rate and size, that re-encode resampled to $8$ fps, and the full recipe, which also rescales to $512{\times}512$ and sets a GOP of $12$. For ReStraV the frame rate is the whole effect. Fitted on \aigvd, it flags $5\%$ of \aigvd's own clips after the plain re-encode, and $72\%$ of them and $57$ to $72\%$ of the other collections after the $8$ fps step, close to the full recipe's rates. WaveRep's head responds to the codec. It flags $24$ to $27\%$ of \dvf's MPEG-4 YouTube clips as shipped and at most $6\%$ after the plain H.264 re-encode, so its over-flagging of \dvf comes from \dvf's encoding. WaveRep's backbone responds mostly to the last step, which adds the square resize (\cref{app:reprocess}). CLIP moves by at most $0.07$ under any step. A real collection's false-positive rate therefore reflects how it was processed as well as what it shows.

Processing every collection alike ends the over-flagging by CLIP and WaveRep's head but not by ReStraV or WaveRep's backbone (\cref{tab:recipefpr}). Here the detectors are also trained and thresholded on recipe clips, so every collection shares one codec, frame rate, and frame size. WaveRep's head then stays at or below $9\%$ on every collection, \dvf included. CLIP's highest rate falls to $13\%$, on \aigvd's real clips. Fitted on \aigvd, ReStraV still flags $43\%$ of MSR-VTT and $12$ to $20\%$ of the other collections, and WaveRep's backbone $35\%$ of MSR-VTT and $29\%$ of Kinetics-400. Resampling MSR-VTT from $3$ to $8$ fps repeats its frames, so the recipe leaves it with repeats. With every collection processed alike, every rate of $15\%$ or more comes from a detector fitted on \aigvd or falls on \aigvd's real clips (\cref{tab:originfpr,tab:recipefpr}).

\subsection{Where a Transfer Loss Falls}
\label{sec:sides}

A cross-benchmark loss can come from either class (\cref{tab:sides} in \cref{app:recipe}). We apply each detector to the other benchmark at its home threshold. A false-positive rate above $5\%$ on the target's real clips then marks a loss on the real side, and a true-positive rate below the rate at home marks a loss on the generated side. CLIP fitted on \gvb loses on both sides. It flags $15\%$ of \aigvd's real clips and catches $42\%$ of its generated ones, against $67\%$ at home. ReStraV fitted on \gvb keeps its rate on \aigvd's real clips ($5\%$) but catches $22\%$ of the generated ones against $92\%$ at home, a loss on the generated side alone. WaveRep's head keeps its false-positive rate within one point of $5\%$ and catches $93$ to $97\%$ in both directions. The side a loss falls on thus depends on the detector and on the direction, so a single cross-benchmark AUC cannot say whether a detector failed on the real collection or on the generators.

\subsection{Distance, and What Restores the Rate}
\label{sec:law}
\label{sec:realonly}

\begin{figure}[t]
  \centering
  \begin{subfigure}[b]{0.52\textwidth}
    \centering
    \includegraphics[width=\linewidth]{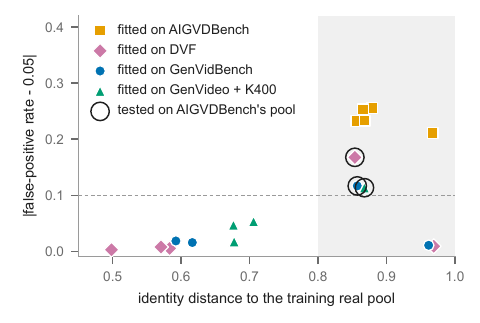}
    \caption{}
    \label{fig:law}
  \end{subfigure}\hfill
  \begin{subfigure}[b]{0.44\textwidth}
    \centering
    \includegraphics[width=\linewidth]{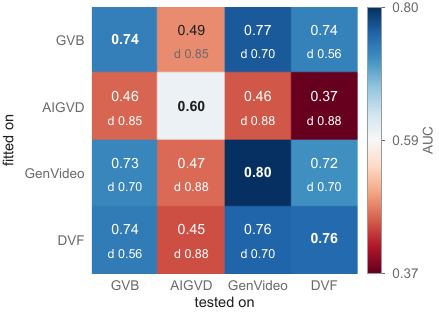}
    \caption{}
    \label{fig:transfer}
  \end{subfigure}
  \caption{\textbf{False-positive departures and transfer losses both occur at large identity distance.} Codec-motion probe under the recipe. (a) One point per training cell and outside collection, plotting the departure $|\mathrm{FPR}-0.05|$ against identity distance. Squares mark \aigvd as the training cell, and open rings mark its real pool as the target. (b) AUC of the probe fitted on the row cell and scored on the column cell, with within-cell LOGO OOD on the diagonal. Red marks low AUC. $d$ is the identity distance between the two cells' pooled real clips (up to $1{,}000$ clips per side, averaged over two subsamples), so it can sit below the per-pool values of \cref{fig:census}. GVB: \gvb; AIGVD: \aigvd.}
  \label{fig:distance}
\end{figure}

Large false-positive departures occur only at large identity distance (\cref{fig:law}). We use the codec-motion probe, the one feature set fitted on all four cells, and take each of the $18$ combinations of a training cell and an outside collection. For each, we plot the \emph{departure} $|\mathrm{FPR}-0.05|$ of the false-positive rate (FPR) from $5\%$ against the identity distance between the cell's real clips and the collection. The eight combinations that depart by more than $0.1$ all lie at distance $0.80$ or more. The eight nearer than $0.80$ depart by at most $0.05$. The remaining two, MSR-VTT seen from \gvb and from \dvf, lie at $0.96$ to $0.97$ and depart by less than $0.02$. Distance ranks the departures only loosely (Spearman rank correlation $0.54$, uncorrected $p{=}0.02$), and the $18$ combinations share four training cells, so they are not independent.

Choosing the threshold on real videos of the target brings the false-positive rate back to about $5\%$, at a cost in true positives that is largest for the detectors with the lowest AUC on the target. We choose it on half of the target's real clips ($231$ to $3{,}000$ clips) and measure the rate on the other half. Both halves come from the same target, so the rate returns by design, and the result is the cost in true positives. Averaged over five seeds, that rate lies between $4.1$ and $6.8\%$ for every detector and every pair of cells we tested (\cref{tab:recal}). Averaged over the two directions of \cref{tab:sides}, ReStraV then catches $22\%$ of the target's generated clips against $39\%$ at its home threshold. CLIP catches $35\%$ against $40\%$, WaveRep's head $93\%$ against $95\%$, and its backbone $99\%$ against $99\%$. Over all twelve pairs of cells, the codec-motion probe catches $9\%$ against $23\%$.

\begin{table}[t]
  \centering
  \footnotesize
  \setlength{\tabcolsep}{4pt}
  \renewcommand{\arraystretch}{1.02}
  \caption{\textbf{Recalibrating on the target restores the false-positive rate and costs true positives.} Rates on held-out clips of the target, five seeds. Home: the threshold set on the training cell. Pairs: ordered pairs of training and target cells. FPR seed means range over pairs, single runs over every pair and seed.}
  \label{tab:recal}
  \fitwidth{%
  \begin{tabular}{@{}lcccccc@{}}
    \toprule
    & & \multicolumn{3}{c}{FPR} & \multicolumn{2}{c}{TPR} \\
    \cmidrule(lr){3-5}\cmidrule(l){6-7}
    Detector & Pairs & Home & Recalibrated, seed means & Recalibrated, single runs & Home & Recalibrated \\
    \midrule
    Codec-motion probe & 12 & $.18$ & $.046$--$.068$ & $.022$--$.121$ & $.23$ & $.09$ \\
    CLIP, PCA-13 & 2 & $.09$ & $.043$--$.052$ & $.021$--$.057$ & $.40$ & $.35$ \\
    ReStraV & 2 & $.09$ & $.048$--$.054$ & $.034$--$.070$ & $.39$ & $.22$ \\
    WaveRep head & 2 & $.05$ & $.041$--$.049$ & $.024$--$.063$ & $.95$ & $.93$ \\
    WaveRep backbone & 2 & $.08$ & $.044$--$.051$ & $.037$--$.060$ & $.99$ & $.99$ \\
    \bottomrule
  \end{tabular}}
\end{table}

Removing identity from the features themselves does not do better. Projecting out the feature directions that separate the training benchmark's real clips from the target's does not beat correlation alignment (CORAL)~\citep{sun2016coral} on mean AUC ($0.797$ against $0.804$, \cref{app:realonly}). The removed directions are specific to each pair of collections.

\section{Do Within-Benchmark Results Replicate and Transfer?}
\label{sec:downstream}

A score \emph{replicates} when the same feature set, re-fitted on a second benchmark, keeps its score there (\cref{sec:transfer}). It \emph{transfers} when a detector fitted on one cell keeps its score on another (\cref{sec:transfertest}).

\subsection{Replication on a Second Benchmark}
\label{sec:transfer}

Of the six feature sets that cleared their \gvb floor by $0.15$ or more, only WaveRep's two replicate on \aigvd (last two columns of \cref{tab:identity}). The test compares each feature set with itself, so it needs no floor on \aigvd. ReStraV, FVMD, and RAFT fell by more than $0.20$. ReStraV went from $0.931$ to $0.673$, and its interval over held-out generators, $[0.422,0.894]$, spans chance. Read through the recipe on both cells, the codec-motion probe fell by $0.13$, against the $-0.230$ of \cref{tab:identity}, which mixes two pipelines. WaveRep's head moved only from $0.996$ to $0.981$, and its backbone stayed at $0.999$. WaveRep's checkpoint was trained on four generators, and two of them, CogVideoX-1.5 and Pyramid-Flow, are among the five \aigvd generators we score, so this replication may partly reflect generators the checkpoint saw in training.

We cannot isolate why the other scores fall. Identity learned from \aigvd's real clips would raise a score there. A cue shared only by the training generators could lower the score on a held-out generator. Processing may also have raised the \gvb score that the drop is measured from, as the repeated frames did for ReStraV (\cref{sec:floors}). The two benchmarks also differ in their generators, in how their clips were processed, and in size (\cref{app:cells}). The floor did not predict which scores replicate. FVMD and ReStraV had the two largest drops, although their floors ($0.574$ and $0.586$) were close to that of WaveRep's head ($0.534$), and WaveRep's backbone replicated at the highest floor of the six, $0.777$.

Putting every clip through the recipe does not make the scores replicate either (lower block of \cref{tab:identity}). With every clip through the recipe, ReStraV rises from $0.601$ on \gvb to $0.783$ on \aigvd, and CLIP falls from $0.919$ to $0.629$. Both WaveRep feature sets stay between $0.97$ and $0.99$, the only scores that replicate both as shipped and under the recipe.

\subsection{Transfer across Four Cells}
\label{sec:transfertest}

Transfer of the codec-motion probe falls below chance only on the pairs that involve \aigvd, which are also the farthest apart (\cref{fig:transfer}). If a detector learns its benchmark's real pool, its score on another benchmark should fall as the two real pools grow further apart. Within each of the four cells the probe reaches $0.60$ to $0.80$. Across cells it ranges from $0.77$ down to $0.37$. Over the $12$ ordered pairs of cells it tracks the distance between their real clips (Spearman $-0.88$). The pairs share four cells, so we test this correlation by permuting the four cells, which gives $p{=}0.04$, the smallest value four cells allow. The six pairs that involve \aigvd have the largest distances ($0.85$ to $0.88$) and score below chance ($0.37$ to $0.49$). Without \aigvd, the remaining six pairs show no trend (Spearman $-0.12$). Fitting on \gvb with both of its real pools transfers to \aigvd no better than fitting with either pool alone ($0.488$ against $0.480$ and $0.496$).

\section{Recommendations, Limitations, and Conclusion}
\label{sec:conclusion}

\para{Recommendations.} A benchmark meant to measure detection should ship at least two independently curated real sources with per-clip source labels, so that a floor can be measured. It should also ship clips of matched length or declare a length filter, and it should declare its encoder settings and frame rate. A detector paper should report three checks on the identity channels: a container probe on the clips it scores, its score with every clip through one shared re-encode such as the recipe, and its floor through its own features on every cell that permits one. The margin over the floor flags scores that may rest on origin, but clearing it did not predict replication here (\cref{sec:transfer}). A paper should therefore also report its score on a second cell, with the false- and true-positive rates at its training threshold, and its false-positive rate on real collections outside its training data. A detector deployed on a new collection should have its threshold set on real videos from that collection (\cref{sec:realonly}).

\para{Release.}
\label{sec:toolkit}
The VidAudit toolkit (\url{https://github.com/KurbanIntelligenceLab/vidaudit}) runs every control of this paper on a manifest of clip paths, labels, and sources (\cref{app:protocol}). It ships with the clip lists of every evaluation set, the census manifests, the probes, the false-positive and transfer analyses, the survey evidence, and the dated construction protocol of the third cell.

\para{Limitations.} The census covers the four benchmarks that ship raw video and that we could download. The probes read three of the cues a detector might use: the container, codec motion, and CLIP appearance. The number of bits spent on each frame, which a CRF encode adapts to content, is not probed, so the fixed pattern is shown to close only the frame-type leak. The recipe also resamples the frame rate and frame size, so a score that falls under it may have lost generation cues as well as origin cues. The length filter is itself a selection that drops short-clip generators. Every score comes from our linear readout, which for WaveRep's head reads the released classifier's frame scores, so the released classifiers may flag different collections. The distance analyses of \cref{fig:distance} use the codec-motion probe alone, whose within-cell scores are modest ($0.60$ to $0.80$), and every large effect in them involves \aigvd's real pool. We score MSR-VTT as \genvideo distributes it, at $3$ fps, so on that collection the census and the false-positive analyses cannot separate what it shows from how it was processed.

\para{Conclusion.} A score on a public AI-generated video benchmark can depend on where its real clips came from as well as on what generated the fakes. Frame types leak until the encoder's frame-type decisions are fixed and the shortest clips set aside. Real collections also separate through motion and appearance, and only one of four corpora ships the two labeled real sources that a floor needs. In deployment, a detector's false-positive rate depends on where a real video came from and how it was processed. By our definition any such shift between real collections is corpus identity, and we split it into processing and content only where a re-encode undoes the processing. Setting the threshold on real videos from the target collection brings that rate back to about $5\%$, at a cost in detected fakes. Identity does not account for every failure, since scores that cleared their floors still failed to replicate and some transfer losses fell on the generated clips. A benchmark score is therefore best read together with its floor, a score on a second cell, and false-positive rates on real video from other collections.

\label{refs-start}
{\small
\bibliography{references}

\begin{thebibliography}{54}
\providecommand{\natexlab}[1]{#1}
\providecommand{\url}[1]{\texttt{#1}}
\expandafter\ifx\csname urlstyle\endcsname\relax
  \providecommand{\doi}[1]{doi: #1}\else
  \providecommand{\doi}{doi: \begingroup \urlstyle{rm}\Url}\fi

\bibitem[Abu-El-Haija et~al.(2016)Abu-El-Haija, Kothari, Lee, Natsev, Toderici,
  Varadarajan, and Vijayanarasimhan]{abuelhaija2016youtube8m}
Sami Abu-El-Haija, Nisarg Kothari, Joonseok Lee, Paul Natsev, George Toderici,
  Balakrishnan Varadarajan, and Sudheendra Vijayanarasimhan.
\newblock {YouTube-8M}: A large-scale video classification benchmark, 2016.
\newblock URL \url{https://arxiv.org/abs/1609.08675}.

\bibitem[Bai et~al.(2025)Bai, Lin, Cao, and Lou]{bai2024aigvdet}
Jianfa Bai, Man Lin, Gang Cao, and Zijie Lou.
\newblock {AI}-generated video detection via spatial-temporal anomaly learning.
\newblock In Zhouchen Lin, Ming-Ming Cheng, Ran He, Kurban Ubul, Wushouer
  Silamu, Hongbin Zha, Jie Zhou, and Cheng-Lin Liu, editors, \emph{Pattern
  Recognition and Computer Vision}, pages 460--470, Singapore, 2025. Springer
  Nature Singapore.
\newblock ISBN 978-981-97-8792-0.

\bibitem[Chandra et~al.(2026)Chandra, Lee, Murtfeldt, Qiu, Karmakar,
  Tanumihardja, Farhat, Caffee, Lee, Choi, Paik, Kim, and
  Etzioni]{chandra2025deepfakeeval}
Nuria~Alina Chandra, Hannah Lee, Ryan Murtfeldt, Lin Qiu, Arnab Karmakar,
  Emmanuel Tanumihardja, Kevin Farhat, Ben Caffee, Changyeon Lee, Jongwook
  Choi, Sejin Paik, Aerin Kim, and Oren Etzioni.
\newblock {Deepfake-Eval-2024}: A multi-modal in-the-wild benchmark of
  deepfakes circulated in 2024.
\newblock In \emph{Proceedings of the IEEE/CVF Conference on Computer Vision
  and Pattern Recognition (CVPR) Workshops}, pages 10668--10678, June 2026.

\bibitem[Chang et~al.(2025)Chang, Liu, Liu, Lyu, Huang, Tao, Wan, Zhang, and
  Qi]{chang2024whatmatters}
Chirui Chang, Jiahui Liu, Zhengzhe Liu, Xiaoyang Lyu, Yi-Hua Huang, Xin Tao,
  Pengfei Wan, Di~Zhang, and Xiaojuan Qi.
\newblock How far are {AI}-generated videos from simulating the {3D} visual
  world: A learned {3D} evaluation approach.
\newblock In \emph{Proceedings of the IEEE/CVF International Conference on
  Computer Vision (ICCV)}, pages 10307--10317, October 2025.

\bibitem[Chen et~al.(2026)Chen, Hong, Huang, Xu, Gu, Li, Lan, Zhu, Zhang, Wang,
  and Li]{chen2024demamba}
Haoxing Chen, Yan Hong, Zizheng Huang, Zhuoer Xu, Zhangxuan Gu, Yaohui Li, Jun
  Lan, Huijia Zhu, Jianfu Zhang, Weiqiang Wang, and Huaxiong Li.
\newblock {DeMamba}: {AI}-generated video detection on million-scale {GenVideo}
  benchmark.
\newblock \emph{Science China Information Sciences}, 69\penalty0 (6):\penalty0
  162103, May 2026.
\newblock ISSN 1869-1919.
\newblock \doi{10.1007/s11432-024-4894-0}.
\newblock URL \url{https://doi.org/10.1007/s11432-024-4894-0}.

\bibitem[Cheng et~al.(2026)Cheng, Yan, Yan, Gan, Zhang, Wang, Peng, and
  Liang]{cheng2026sanity}
Jikang Cheng, Renye Yan, Zhiyuan Yan, Yaozhong Gan, Xueyi Zhang, Zhongyuan
  Wang, Wei Peng, and Ling Liang.
\newblock A sanity check for multi-in-domain face forgery detection in the real
  world.
\newblock In \emph{Proceedings of the IEEE/CVF Conference on Computer Vision
  and Pattern Recognition (CVPR)}, pages 21306--21315, June 2026.

\bibitem[Choe et~al.(2020)Choe, Oh, Lee, Chun, Akata, and
  Shim]{choe2020evaluating}
Junsuk Choe, Seong~Joon Oh, Seungho Lee, Sanghyuk Chun, Zeynep Akata, and
  Hyunjung Shim.
\newblock Evaluating weakly supervised object localization methods right.
\newblock In \emph{Proceedings of the IEEE/CVF Conference on Computer Vision
  and Pattern Recognition (CVPR)}, June 2020.

\bibitem[Cokelaer and Hasch(2017)]{cokelaer2017spectrum}
Thomas Cokelaer and Juergen Hasch.
\newblock `{Spectrum}': Spectral analysis in {Python}.
\newblock \emph{The Journal of Open Source Software}, 2\penalty0 (18):\penalty0
  348, 2017.
\newblock \doi{10.21105/joss.00348}.

\bibitem[Corvi et~al.(2025)Corvi, Cozzolino, Prashnani, De~Mello, Nagano, and
  Verdoliva]{corvi2025waverep}
Riccardo Corvi, Davide Cozzolino, Ekta Prashnani, Shalini De~Mello, Koki
  Nagano, and Luisa Verdoliva.
\newblock Seeing what matters: Generalizable {AI}-generated video detection
  with forensic-oriented augmentation.
\newblock In D.~Belgrave, C.~Zhang, H.~Lin, R.~Pascanu, P.~Koniusz,
  M.~Ghassemi, and N.~Chen, editors, \emph{Advances in Neural Information
  Processing Systems}, volume~38, pages 26418--26446. Curran Associates, Inc.,
  2025.
\newblock URL
  \url{https://proceedings.neurips.cc/paper_files/paper/2025/file/25e92e33ac8c35fd49f394c37f21b6da-Paper-Conference.pdf}.

\bibitem[Du et~al.(2026)Du, Chen, Li, Zhang, Ouyang, Zhang, and
  Liu]{du2026g2vd}
Meng Du, Hongchang Chen, Ran Li, Junjie Zhang, Qi~Ouyang, Shibo Zhang, and
  Shuxin Liu.
\newblock {G2VD}: Generalizable {AI}-generated video detection via
  counterfactual intervention and causal disentanglement.
\newblock arXiv:2607.04607, 2026.

\bibitem[Fang et~al.(2026)Fang, Wei, Wang, Zhang, Yu, and
  Zhou]{fang2026realdatalies}
Ziyuan Fang, Tianyi Wei, Guanjie Wang, Weiming Zhang, Nenghai Yu, and Wenbo
  Zhou.
\newblock Real data lies: Unveiling and closing the quality shortcut in
  generalizable {AI}-generated video detection.
\newblock In \emph{Proceedings of the 43rd International Conference on Machine
  Learning}, volume 306 of \emph{Proceedings of Machine Learning Research},
  pages 29399--29416. PMLR, Jul 2026.

\bibitem[{FFmpeg Developers}()]{ffmpeg}
{FFmpeg Developers}.
\newblock {FFmpeg}.
\newblock \url{https://ffmpeg.org/}.
\newblock Version 7.1.1.

\bibitem[Grommelt et~al.(2025)Grommelt, Weiss, Pfreundt, and
  Keuper]{grommelt2024fakejpeg}
Patrick Grommelt, Louis Weiss, Franz-Josef Pfreundt, and Janis Keuper.
\newblock Fake or {JPEG}? {R}evealing common biases in generated image
  detection datasets.
\newblock In \emph{Computer Vision -- ECCV 2024 Workshops}, pages 80--95.
  Springer, Cham, 2025.
\newblock \doi{10.1007/978-3-031-92089-9_6}.

\bibitem[Guillaro et~al.(2025)Guillaro, Zingarini, Usman, Sud, Cozzolino, and
  Verdoliva]{guillaro2025bfree}
Fabrizio Guillaro, Giada Zingarini, Ben Usman, Avneesh Sud, Davide Cozzolino,
  and Luisa Verdoliva.
\newblock A bias-free training paradigm for more general {AI}-generated image
  detection.
\newblock In \emph{Proceedings of the IEEE/CVF Conference on Computer Vision
  and Pattern Recognition (CVPR)}, pages 18685--18694, June 2025.

\bibitem[He et~al.(2024)He, Zhu, Li, Wang, and Li]{he2024lgtd}
Peisong He, Leyao Zhu, Jiaxing Li, Shiqi Wang, and Haoliang Li.
\newblock Exposing {AI}-generated videos: A benchmark dataset and a
  local-and-global temporal defect based detection method, 2024.
\newblock URL \url{https://arxiv.org/abs/2405.04133}.

\bibitem[Intern\`{o} et~al.(2025)Intern\`{o}, Geirhos, Olhofer, Liu, Hammer,
  and Klindt]{restrav2024}
Christian Intern\`{o}, Robert Geirhos, Markus Olhofer, Sunny Liu, Barbara
  Hammer, and David Klindt.
\newblock {AI}-generated video detection via perceptual straightening.
\newblock In D.~Belgrave, C.~Zhang, H.~Lin, R.~Pascanu, P.~Koniusz,
  M.~Ghassemi, and N.~Chen, editors, \emph{Advances in Neural Information
  Processing Systems}, volume~38, pages 20672--20705. Curran Associates, Inc.,
  2025.
\newblock URL
  \url{https://proceedings.neurips.cc/paper_files/paper/2025/file/1d9a43752c2819e03967c5c1b708169c-Paper-Conference.pdf}.

\bibitem[Kay et~al.(2017)Kay, Carreira, Simonyan, Zhang, Hillier,
  Vijayanarasimhan, Viola, Green, Back, Natsev, Suleyman, and
  Zisserman]{kay2017kinetics}
Will Kay, Joao Carreira, Karen Simonyan, Brian Zhang, Chloe Hillier, Sudheendra
  Vijayanarasimhan, Fabio Viola, Tim Green, Trevor Back, Paul Natsev, Mustafa
  Suleyman, and Andrew Zisserman.
\newblock The {Kinetics} human action video dataset, 2017.
\newblock URL \url{https://arxiv.org/abs/1705.06950}.

\bibitem[Kundu et~al.(2025)Kundu, Xiong, Mohanty, Balachandran, and
  Roy-Chowdhury]{kundu2025unite}
Rohit Kundu, Hao Xiong, Vishal Mohanty, Athula Balachandran, and Amit~K.
  Roy-Chowdhury.
\newblock Towards a universal synthetic video detector: From face or background
  manipulations to fully {AI}-generated content.
\newblock In \emph{Proceedings of the IEEE/CVF Conference on Computer Vision
  and Pattern Recognition (CVPR)}, pages 28050--28060, June 2025.

\bibitem[Ku{\'z}nicka-B{\l}aszkowska(2026)]{kuznicka2026deepfakes}
Dominika Ku{\'z}nicka-B{\l}aszkowska.
\newblock Deepfakes, financial stability, and {EU} regulation: Navigating the
  risks of synthetic media.
\newblock \emph{Capital Markets Law Journal}, 21\penalty0 (2):\penalty0
  kmag008, April 2026.
\newblock \doi{10.1093/cmlj/kmag008}.
\newblock URL \url{https://doi.org/10.1093/cmlj/kmag008}.

\bibitem[Layton et~al.(2024)Layton, Tucker, Olszewski, Warren, Butler, and
  Traynor]{layton2024sok}
Seth Layton, Tyler Tucker, Daniel Olszewski, Kevin Warren, Kevin Butler, and
  Patrick Traynor.
\newblock {SoK}: The good, the bad, and the unbalanced: Measuring structural
  limitations of deepfake media datasets.
\newblock In \emph{Proceedings of the 33rd USENIX Security Symposium (USENIX
  Security 24)}, pages 1027--1044, Philadelphia, PA, 2024. USENIX Association.
\newblock ISBN 978-1-939133-44-1.
\newblock URL
  \url{https://www.usenix.org/conference/usenixsecurity24/presentation/layton}.

\bibitem[Li et~al.(2026)Li, Jiang, Zhao, Zhou, Mo, Gao, Yang, Shan, Wu, and
  Su]{li2026nativescale}
Zhengcen Li, Chenyang Jiang, Hang Zhao, Shiyang Zhou, Yunyang Mo, Feng Gao, Fan
  Yang, Qiben Shan, Shaocong Wu, and Jingyong Su.
\newblock Preserving forgery artifacts: {AI}-generated video detection at
  native scale.
\newblock In C.~Vondrick, B.~Hariharan, C.~Raffel, L.~Pinto, D.~Yang, and
  A.~Faust, editors, \emph{International Conference on Learning
  Representations}, pages 19429--19455, 2026.
\newblock URL
  \url{https://proceedings.iclr.cc/paper_files/paper/2026/file/20e45668fefa793bd9f2edf19be12c4b-Paper-Conference.pdf}.

\bibitem[Liu et~al.(2024)Liu, Qu, Yan, Zeng, Wang, and Liao]{liu2024fvmd}
Jiahe Liu, Youran Qu, Qi~Yan, Xiaohui Zeng, Lele Wang, and Renjie Liao.
\newblock Fr\'echet video motion distance: A metric for evaluating motion
  consistency in videos, 2024.
\newblock URL \url{https://arxiv.org/abs/2407.16124}.

\bibitem[Liu and He(2025)]{liu2025decade}
Zhuang Liu and Kaiming He.
\newblock A decade's battle on dataset bias: Are we there yet?
\newblock In \emph{International Conference on Learning Representations
  (ICLR)}, 2025.

\bibitem[Ma et~al.(2025)Ma, Yan, Guo, Liao, Yu, and Zhou]{maaz2024decof}
Long Ma, Zhiyuan Yan, Qinglang Guo, Yong Liao, Haiyang Yu, and Pengyuan Zhou.
\newblock Detecting {AI}-generated video via frame consistency.
\newblock In \emph{2025 IEEE International Conference on Multimedia and Expo
  (ICME)}, pages 1--6, 2025.
\newblock \doi{10.1109/ICME59968.2025.11210049}.

\bibitem[Ma et~al.(2026)Ma, Xue, Wang, Yan, Xu, Jiang, Yu, Liao, and
  Bi]{aigvdbench2024}
Long Ma, Zihao Xue, Yan Wang, Zhiyuan Yan, Jin Xu, Xiaorui Jiang, Haiyang Yu,
  Yong Liao, and Zhen Bi.
\newblock Your one-stop solution for {AI}-generated video detection.
\newblock In \emph{Proceedings of the IEEE/CVF Conference on Computer Vision
  and Pattern Recognition (CVPR)}, pages 4458--4470, June 2026.

\bibitem[Michels et~al.(2026)Michels, Jorissen, and
  Michiels]{michels2026motionbias}
Joren Michels, Lode Jorissen, and Nick Michiels.
\newblock Dataset biases and shortcut learning in motion-based {AI}-generated
  video detection.
\newblock arXiv:2607.00948, 2026.

\bibitem[Ni et~al.(2022)Ni, Peng, Chen, Zhang, Meng, Fu, Xiang, and
  Ling]{ni2022xclip}
Bolin Ni, Houwen Peng, Minghao Chen, Songyang Zhang, Gaofeng Meng, Jianlong Fu,
  Shiming Xiang, and Haibin Ling.
\newblock Expanding language-image pretrained models for general video
  recognition.
\newblock In \emph{Computer Vision -- ECCV 2022}, pages 1--18. Springer Nature
  Switzerland, 2022.
\newblock \doi{10.1007/978-3-031-19772-7_1}.

\bibitem[Ni et~al.(2026)Ni, Yan, Huang, Yuan, Tang, Hu, Chen, and
  Wang]{ni2025genvidbenchchallengingbenchmarkdetecting}
Zhenliang Ni, Qiangyu Yan, Mouxiao Huang, Tianning Yuan, Yehui Tang, Hailin Hu,
  Xinghao Chen, and Yunhe Wang.
\newblock {GenVidBench}: A 6-million benchmark for {AI}-generated video
  detection.
\newblock \emph{Proceedings of the AAAI Conference on Artificial Intelligence},
  40\penalty0 (18):\penalty0 15582--15590, 2026.
\newblock \doi{10.1609/aaai.v40i18.38587}.
\newblock URL \url{https://ojs.aaai.org/index.php/AAAI/article/view/38587}.

\bibitem[Oquab et~al.(2024)Oquab, Darcet, Moutakanni, Vo, Szafraniec, Khalidov,
  Fernandez, Haziza, Massa, El-Nouby, Assran, Ballas, Galuba, Howes, Huang, Li,
  Misra, Rabbat, Sharma, Synnaeve, Xu, Jegou, Mairal, Labatut, Joulin, and
  Bojanowski]{oquab2024dinov2}
Maxime Oquab, Timoth{\'e}e Darcet, Th{\'e}o Moutakanni, Huy~V. Vo, Marc
  Szafraniec, Vasil Khalidov, Pierre Fernandez, Daniel Haziza, Francisco Massa,
  Alaaeldin El-Nouby, Mido Assran, Nicolas Ballas, Wojciech Galuba, Russell
  Howes, Po-Yao Huang, Shang-Wen Li, Ishan Misra, Michael Rabbat, Vasu Sharma,
  Gabriel Synnaeve, Hu~Xu, Herve Jegou, Julien Mairal, Patrick Labatut, Armand
  Joulin, and Piotr Bojanowski.
\newblock {DINO}v2: Learning robust visual features without supervision.
\newblock \emph{Transactions on Machine Learning Research}, 2024.
\newblock ISSN 2835-8856.
\newblock URL \url{https://openreview.net/forum?id=a68SUt6zFt}.
\newblock Featured Certification.

\bibitem[Park et~al.(2026)Park, Yang, Yi, Zheng, Muaz, Shen, Han, Shan, and
  Qiu]{park2025vidguard}
Kyoungjun Park, Yifan Yang, Juheon Yi, Shicheng Zheng, Muhammad Muaz, Yifei
  Shen, Dongqi Han, Caihua Shan, and Lili Qiu.
\newblock {VidGuard-R1}: {AI}-generated video detection and explanation via
  reasoning {MLLMs} and {RL}.
\newblock In C.~Vondrick, B.~Hariharan, C.~Raffel, L.~Pinto, D.~Yang, and
  A.~Faust, editors, \emph{International Conference on Learning
  Representations}, pages 108034--108058, 2026.
\newblock URL
  \url{https://proceedings.iclr.cc/paper_files/paper/2026/file/b00ef390dcd5f147fd7c5c2bb35f09be-Paper-Conference.pdf}.

\bibitem[Radford et~al.(2021)Radford, Kim, Hallacy, Ramesh, Goh, Agarwal,
  Sastry, Askell, Mishkin, Clark, Krueger, and Sutskever]{radford2021clip}
Alec Radford, Jong~Wook Kim, Chris Hallacy, Aditya Ramesh, Gabriel Goh,
  Sandhini Agarwal, Girish Sastry, Amanda Askell, Pamela Mishkin, Jack Clark,
  Gretchen Krueger, and Ilya Sutskever.
\newblock Learning transferable visual models from natural language
  supervision.
\newblock In Marina Meila and Tong Zhang, editors, \emph{Proceedings of the
  38th International Conference on Machine Learning}, volume 139 of
  \emph{Proceedings of Machine Learning Research}, pages 8748--8763. PMLR,
  18--24 Jul 2021.
\newblock URL \url{https://proceedings.mlr.press/v139/radford21a.html}.

\bibitem[Ramkumar(2026)]{ramkumar2026}
Amrith Ramkumar.
\newblock {AI} deepfakes are getting weirder and harder to spot in the
  midterms.
\newblock \emph{The Wall Street Journal}, 2026.
\newblock URL
  \url{https://www.wsj.com/politics/elections/ai-deepfakes-are-getting-weirder-and-harder-to-spot-in-the-midterms-88b4f7ad}.

\bibitem[Smeu et~al.(2025)Smeu, Boldisor, Oneata, and
  Oneata]{smeu2025shortcuts}
Stefan Smeu, Dragos-Alexandru Boldisor, Dan Oneata, and Elisabeta Oneata.
\newblock Circumventing shortcuts in audio-visual deepfake detection datasets
  with unsupervised learning.
\newblock In \emph{Proceedings of the IEEE/CVF Conference on Computer Vision
  and Pattern Recognition (CVPR)}, pages 18815--18825, June 2025.

\bibitem[Song et~al.(2024)Song, Guo, Zhang, Li, Bai, Liu, Zhai, and
  Liu]{song2024mmdet}
Xiufeng Song, Xiao Guo, Jiache Zhang, Qirui Li, Lei Bai, Xiaoming Liu, Guangtao
  Zhai, and Xiaohong Liu.
\newblock On learning multi-modal forgery representation for diffusion
  generated video detection.
\newblock In A.~Globerson, L.~Mackey, D.~Belgrave, A.~Fan, U.~Paquet,
  J.~Tomczak, and C.~Zhang, editors, \emph{Advances in Neural Information
  Processing Systems}, volume~37, pages 122054--122077. Curran Associates,
  Inc., 2024.
\newblock \doi{10.52202/079017-3878}.
\newblock URL
  \url{https://proceedings.neurips.cc/paper_files/paper/2024/file/dccbeb7a8df3065c4646928985edf435-Paper-Conference.pdf}.

\bibitem[Sun et~al.(2016)Sun, Feng, and Saenko]{sun2016coral}
Baochen Sun, Jiashi Feng, and Kate Saenko.
\newblock Return of frustratingly easy domain adaptation.
\newblock \emph{Proceedings of the AAAI Conference on Artificial Intelligence},
  30\penalty0 (1), 2016.
\newblock \doi{10.1609/aaai.v30i1.10306}.
\newblock URL \url{https://ojs.aaai.org/index.php/AAAI/article/view/10306}.

\bibitem[Sundara~Rajan et~al.(2025)Sundara~Rajan, Ojha, Schloesser, and
  Lee]{rajan2025aligned}
Anirudh Sundara~Rajan, Utkarsh Ojha, Jedidiah Schloesser, and Yong~Jae Lee.
\newblock Aligned datasets improve detection of latent diffusion-generated
  images.
\newblock In Y.~Yue, A.~Garg, N.~Peng, F.~Sha, and R.~Yu, editors,
  \emph{International Conference on Learning Representations}, pages
  63304--63322, 2025.
\newblock URL
  \url{https://proceedings.iclr.cc/paper_files/paper/2025/file/9ead108421b202494d01b5060d12aa34-Paper-Conference.pdf}.

\bibitem[Tan et~al.(2026)Tan, Lan, Shi, Tan, Yu, Zhu, Wang, Wan, and
  Lei]{tan2026videoveritas}
Hao Tan, Jun Lan, Senyuan Shi, Zichang Tan, Zijian Yu, Huijia Zhu, Weiqiang
  Wang, Jun Wan, and Zhen Lei.
\newblock {VideoVeritas}: {AI}-generated video detection via perception pretext
  reinforcement learning, 2026.
\newblock URL \url{https://arxiv.org/abs/2602.08828}.

\bibitem[Teed and Deng(2020)]{teed2020raft}
Zachary Teed and Jia Deng.
\newblock {RAFT}: Recurrent all-pairs field transforms for optical flow.
\newblock In Andrea Vedaldi, Horst Bischof, Thomas Brox, and Jan-Michael Frahm,
  editors, \emph{Computer Vision -- ECCV 2020}, pages 402--419, Cham, 2020.
  Springer International Publishing.
\newblock ISBN 978-3-030-58536-5.

\bibitem[Torralba and Efros(2011)]{torralba2011unbiased}
Antonio Torralba and Alexei~A. Efros.
\newblock Unbiased look at dataset bias.
\newblock In \emph{Proceedings of the IEEE Conference on Computer Vision and
  Pattern Recognition (CVPR)}, pages 1521--1528, 2011.
\newblock \doi{10.1109/CVPR.2011.5995347}.

\bibitem[Vaccari and Chadwick(2020)]{vaccari2020deepfakes}
Cristian Vaccari and Andrew Chadwick.
\newblock Deepfakes and disinformation: Exploring the impact of synthetic
  political video on deception, uncertainty, and trust in news.
\newblock \emph{Social Media + Society}, 6\penalty0 (1), 2020.
\newblock \doi{10.1177/2056305120903408}.
\newblock URL \url{https://doi.org/10.1177/2056305120903408}.

\bibitem[V{\'a}zquez-Pad{\'i}n et~al.(2012)V{\'a}zquez-Pad{\'i}n, Fontani,
  Bianchi, Comesa{\~n}a, Piva, and Barni]{vazquezpadin2012gop}
David V{\'a}zquez-Pad{\'i}n, Marco Fontani, Tiziano Bianchi, Pedro
  Comesa{\~n}a, Alessandro Piva, and Mauro Barni.
\newblock Detection of video double encoding with {GOP} size estimation.
\newblock In \emph{IEEE International Workshop on Information Forensics and
  Security (WIFS)}, pages 151--156, 2012.
\newblock \doi{10.1109/WIFS.2012.6412641}.

\bibitem[{VideoLAN}()]{x264}
{VideoLAN}.
\newblock x264, an {H.264/AVC} encoder.
\newblock \url{https://www.videolan.org/developers/x264.html}.

\bibitem[Wang et~al.(2026{\natexlab{a}})Wang, Shen, Lin, Yang, Zhang, and
  Wang]{zhao2024cmta}
Hang Wang, Chao Shen, Chenhao Lin, Minghui Yang, Lei Zhang, and Cong Wang.
\newblock {CMTA}: Leveraging cross-modal temporal artifacts for generalizable
  {AI}-generated video detection, 2026{\natexlab{a}}.
\newblock URL \url{https://arxiv.org/abs/2605.00630}.

\bibitem[Wang et~al.(2026{\natexlab{b}})Wang, Shen, Zhang, and
  Cheng]{xu2024atss}
Hang Wang, Chao Shen, Lei Zhang, and Zhi-Qi Cheng.
\newblock {ATSS}: Detecting {AI}-generated videos via anomalous temporal
  self-similarity, 2026{\natexlab{b}}.
\newblock URL \url{https://arxiv.org/abs/2604.04029}.

\bibitem[Wang and Farid(2006)]{wang2006double}
Weihong Wang and Hany Farid.
\newblock Exposing digital forgeries in video by detecting double {MPEG}
  compression.
\newblock In \emph{Proceedings of the 8th Workshop on Multimedia and Security},
  2006.
\newblock \doi{10.1145/1161366.1161375}.

\bibitem[Wang et~al.(2024{\natexlab{a}})Wang, Yang, Tuo, He, Zhu, Fu, and
  Liu]{wang2023videofactory}
Wenjing Wang, Huan Yang, Zixi Tuo, Huiguo He, Junchen Zhu, Jianlong Fu, and
  Jiaying Liu.
\newblock Swap attention in spatiotemporal diffusions for text-to-video
  generation, 2024{\natexlab{a}}.
\newblock URL \url{https://arxiv.org/abs/2305.10874}.

\bibitem[Wang et~al.(2024{\natexlab{b}})Wang, He, Li, Li, Yu, Ma, Li, Chen,
  Chen, Wang, Luo, Liu, Wang, Wang, and Qiao]{wang2024internvid}
Yi~Wang, Yinan He, Yizhuo Li, Kunchang Li, Jiashuo Yu, Xin Ma, Xinhao Li, Guo
  Chen, Xinyuan Chen, Yaohui Wang, Ping Luo, Ziwei Liu, Yali Wang, Limin Wang,
  and Yu~Qiao.
\newblock {InternVid}: A large-scale video-text dataset for multimodal
  understanding and generation.
\newblock In B.~Kim, Y.~Yue, S.~Chaudhuri, K.~Fragkiadaki, M.~Khan, and Y.~Sun,
  editors, \emph{International Conference on Learning Representations}, pages
  42055--42079, 2024{\natexlab{b}}.
\newblock URL
  \url{https://proceedings.iclr.cc/paper_files/paper/2024/file/b7bfab38ed694b43e8c20c14f6c0e900-Paper-Conference.pdf}.

\bibitem[Wang et~al.(2026{\natexlab{c}})Wang, Liu, and Sun]{wang2025robustsora}
Zhuo Wang, Xiliang Liu, and Ligang Sun.
\newblock {RobustSora}: De-watermarked benchmark for robust {AI}-generated
  video detection, 2026{\natexlab{c}}.
\newblock URL \url{https://arxiv.org/abs/2512.10248}.

\bibitem[Welch(1967)]{welch1967psd}
P.~Welch.
\newblock The use of fast {Fourier} transform for the estimation of power
  spectra: A method based on time averaging over short, modified periodograms.
\newblock \emph{IEEE Transactions on Audio and Electroacoustics}, 15\penalty0
  (2):\penalty0 70--73, 1967.
\newblock \doi{10.1109/TAU.1967.1161901}.

\bibitem[Xu et~al.(2016)Xu, Mei, Yao, and Rui]{xu2016msrvtt}
Jun Xu, Tao Mei, Ting Yao, and Yong Rui.
\newblock {MSR-VTT}: A large video description dataset for bridging video and
  language.
\newblock In \emph{Proceedings of the IEEE Conference on Computer Vision and
  Pattern Recognition (CVPR)}, June 2016.

\bibitem[Yang et~al.(2024)Yang, Huang, Lu, Han, Zhang, Gao, Hu, and
  Zhao]{yang2024vript}
Dongjie Yang, Suyuan Huang, Chengqiang Lu, Xiaodong Han, Haoxin Zhang, Yan Gao,
  Yao Hu, and Hai Zhao.
\newblock {Vript}: A video is worth thousands of words.
\newblock In A.~Globerson, L.~Mackey, D.~Belgrave, A.~Fan, U.~Paquet,
  J.~Tomczak, and C.~Zhang, editors, \emph{Advances in Neural Information
  Processing Systems}, volume~37, pages 57240--57261. Curran Associates, Inc.,
  2024.
\newblock \doi{10.52202/079017-1824}.
\newblock URL
  \url{https://proceedings.neurips.cc/paper_files/paper/2024/file/6903a5aaece71b76623245fc6e32f01b-Paper-Datasets_and_Benchmarks_Track.pdf}.

\bibitem[Zhang et~al.(2025)Zhang, Lian, Yang, Li, Pang, Liu, Han, Li, and
  Tan]{zhang2025nsgvd}
Shuhai Zhang, ZiHao Lian, Jiahao Yang, Daiyuan Li, Guoxuan Pang, Feng Liu,
  Bo~Han, Shutao Li, and Mingkui Tan.
\newblock Physics-driven spatiotemporal modeling for {AI}-generated video
  detection.
\newblock In D.~Belgrave, C.~Zhang, H.~Lin, R.~Pascanu, P.~Koniusz,
  M.~Ghassemi, and N.~Chen, editors, \emph{Advances in Neural Information
  Processing Systems}, volume~38, pages 174683--174730. Curran Associates,
  Inc., 2025.
\newblock URL
  \url{https://proceedings.neurips.cc/paper_files/paper/2025/file/ff7c9c90030f1eb3cbf5c81b6fbd9a05-Paper-Conference.pdf}.

\bibitem[Zheng et~al.(2025)Zheng, Suo, Lin, Zhao, Yang, Liu, Yang, Wang, and
  Shen]{ma2024d3}
Chende Zheng, Ruiqi Suo, Chenhao Lin, Zhengyu Zhao, Le~Yang, Shuai Liu, Minghui
  Yang, Cong Wang, and Chao Shen.
\newblock {D3}: Training-free {AI}-generated video detection using second-order
  features.
\newblock In \emph{Proceedings of the IEEE/CVF International Conference on
  Computer Vision (ICCV)}, pages 12852--12862, October 2025.

\bibitem[Zheng et~al.(2023)Zheng, Harley, Shen, Wetzstein, and
  Guibas]{zheng2023pointodyssey}
Yang Zheng, Adam~W. Harley, Bokui Shen, Gordon Wetzstein, and Leonidas~J.
  Guibas.
\newblock {PointOdyssey}: A large-scale synthetic dataset for long-term point
  tracking.
\newblock In \emph{Proceedings of the IEEE/CVF International Conference on
  Computer Vision (ICCV)}, pages 19855--19865, October 2023.

\end{thebibliography}
}
\label{refs-end}
\clearpage
\appendix
\crefalias{section}{appendix}
\crefalias{subsection}{subappendix}
\raggedbottom
\renewcommand{\topfraction}{0.7}
\renewcommand{\bottomfraction}{0.5}
\renewcommand{\textfraction}{0.3}
\renewcommand{\floatpagefraction}{0.85}
\setcounter{topnumber}{2}
\setcounter{bottomnumber}{1}
\setcounter{totalnumber}{3}
\makeatletter\setlength{\@fptop}{0pt}\makeatother
\renewcommand{\arraystretch}{1.02}

\noindent \cref{app:probes} defines the three probes. \cref{app:ci} gives the readout and the interval procedures behind every floor and score. \cref{app:cells} defines the cells and their pipelines, and \cref{app:protocol} lists the controls. \cref{app:container} completes the container-probe grid, \cref{app:censusdetail} gives the census sources, and \cref{app:foldci} gives the fold-level intervals of the replication column. \cref{app:gates} sweeps the margin cut and the length filter. \cref{app:survey} gives the survey of $20$ papers. \cref{app:adaptive} gives the two adversaries and the paired test, \cref{app:reprocess} the re-processing controls, \cref{app:recipe} the detectors under the recipe, \cref{app:realonly} the identity-removal experiment, and \cref{app:harness} per-detector readout details behind \cref{tab:identity}.

\section{Probes and Feature Definitions}
\label{app:probes}

The three probes of \cref{sec:probes} share one readout with every detector. This section fixes their inputs.

\para{Container probe.} In the census and on every cell after the recipe, frame types are the picture types the decoder reports for each frame, which the bitstream declares in its frame headers. On the \gvb cell under its native pipeline, they were read from the reference-frame index that FFmpeg~\citep{ffmpeg} exports with each block's motion vector. A frame is labeled I when no block references another frame, B when any block references a future frame, and P otherwise. The three-feature probe is the shared readout on the counts $(n_I, n_P, n_B)$ of one clip. The one-feature probe reads the total frame count $n_\text{total}$ alone. Under the length filter both probes read the prefix that ends at the $K$th non-I-frame, which is the part of the clip a motion detector sees after the filter.

\para{Codec-motion probe.} The thirteen features summarize the motion vectors of the first $K$ non-I-frames, four by how much motion there is (Tier~0) and nine by how it is structured over time (Tier~1). The vectors, one per $16{\times}16$ macroblock, are read from the motion-vector side data that FFmpeg's H.264 decoder exports, and resampled onto a $32{\times}32$ grid per frame. Let $m_t^{(i,j)}$ be the motion-vector magnitude at grid position $(i,j)$ in frame $t$, let $N_\text{MB}$ be the number of positions per frame, and let $\bar m_t = \frac{1}{N_\text{MB}}\sum_{i,j} m_t^{(i,j)}$ be the mean magnitude of frame $t$. Tier~0 holds four clip-level scalars that measure how much motion there is. T0-1 is the mean of $\bar m_t$ over the $K$ frames. T0-2 is the sparsity, the fraction of positions with zero magnitude. T0-3 is the variance $\mathrm{Var}_t(\bar m_t)$. T0-4 is the mean absolute temporal difference $|\bar m_t - \bar m_{t-1}|$.

Tier~1 measures how the motion is structured over time. The grid is split into a $4{\times}4$ arrangement of $16$ regions. For each region, $r(t)$ is the mean of $m_t^{(i,j)}$ over the region in frame $t$. Its power spectral density $S_r(f)$ is estimated with Welch's periodogram~\citep{welch1967psd} for series of $32$ frames or more. Shorter series, as at $K{=}12$, use a fourth-order autoregressive model fitted by Burg's method as implemented in Spectrum~\citep{cokelaer2017spectrum}. The maximum autocorrelation lag is $L{=}6$. Six per-region statistics follow. T1-1 is the spectral slope $\beta$ of a log-log fit $\log S_r(f) = C - \beta \log f$. T1-2 is the spectral flatness, the ratio of the geometric to the arithmetic mean of $S_r(f)$. T1-3 is the decay time $\tau_0$ of the autocorrelation $R(\tau)\approx\exp(-\tau/\tau_0)$. T1-4 is the first lag at which $R(\tau)$ crosses zero. T1-5 is the excess kurtosis of the acceleration $a_t = r(t) - 2r(t-1) + r(t-2)$, and T1-6 is its skewness. The thirteen features are the four Tier~0 scalars, the median over the $16$ regions of T1-1 to T1-6, and the interquartile range over the regions of T1-1 to T1-3. Every Tier~1 feature is unchanged when $r(t)$ is multiplied by a positive constant, because each is a ratio of quantities that scale together. Tier~0 carries that scale, except sparsity, which counts zero positions. The spectral estimate, the maximum lag, and the per-region spread all depend on how many frames a clip contributes, which is why the length filter fixes that number at $K{=}12$. The probe runs on one CPU core at $14.2$\,ms per clip after decoding.

\para{Appearance probe.} Each decoded frame of the clip, in the pipeline the table names, is embedded with CLIP ViT-B/32 and the embeddings are mean-pooled into one $512$-dimensional vector per clip. A PCA to $13$ components, fitted on the training fold of each split, gives the probe the same dimension as the two others. \cref{sec:floor} reports the floor at both dimensions.

\section{Readout, Intervals, and Seed Spread}
\label{app:ci}

\para{Readout.} Every score and every floor in this paper comes from one readout, re-fitted for every compared feature set. Features are median-imputed and standardized per feature on the training fold, and an $L_2$-regularized logistic regression (L2-LR) with balanced class weights and regularization strength $C{=}1$ is fitted. The $80/20$ split score fits one stratified $80/20$ split over all generators at once. Leave-one-generator-out (LOGO) scores are means over the held-out generators. LOGO in-distribution (ID), reported only in the appendix tables, scores the same test clips as LOGO OOD with a readout trained on real clips and on a separate training split of the held-out generator. The identity floor is the AUC of the same readout fitted to tell the two real pools apart, on its own stratified split. All splits are deterministic at seed $42$.

\para{Intervals.} Two interval procedures appear in the paper. A test-set bootstrap ($1{,}000$ resamples of the held-out predictions, percentile method, seed $42$) accompanies the census floor quoted in \cref{sec:floors}. A fold-level bootstrap ($1{,}000$ resamples of the per-generator LOGO scores) accompanies every replication score on \aigvd (\cref{app:foldci}) and every LOGO OOD in \cref{tab:supp_headline_ci}. Seed spread is the standard deviation over the seeds $\{42,0,7,17,99\}$, each with a fresh stratified $80/20$ split, fresh LOGO folds, and a fresh floor split, with the feature set and the estimator held fixed. The seed-$42$ values quoted in the main text sit within two standard deviations of the multi-seed means in every row of \cref{tab:supp_seeds_baselines}, so they are not favorable selections. WaveRep's checkpoint is fixed and only the readout on its three LLR summaries is fitted, so its spread comes from the splits alone.

\begin{table}[htb]
\centering
\footnotesize
\setlength{\tabcolsep}{8pt}
\caption{\textbf{Multi-seed mean $\pm$ std for the fitted rows of \cref{tab:identity}.} Seeds $\{0,7,17,42,99\}$, \gvb cell, shared readout. The $80/20$ split column fits one split over all generators. The PCA-$13$ rows fit the PCA on the training fold of each split.}
\label{tab:supp_seeds_baselines}
{%
\begin{tabular}{@{}lcccc@{}}
\toprule
Feature source & $80/20$ split & LOGO ID & LOGO OOD & Floor (\rvr) \\
\midrule
Codec-motion probe ($13$-d)  & $0.862 \pm 0.007$ & $0.907 \pm 0.002$ & $0.828 \pm 0.003$ & $0.628 \pm 0.017$ \\
FVMD full ($1024$-d)         & $0.908 \pm 0.003$ & $0.926 \pm 0.004$ & $0.886 \pm 0.004$ & $0.572 \pm 0.008$ \\
FVMD PCA-$13$                & $0.869 \pm 0.005$ & $0.877 \pm 0.005$ & $0.860 \pm 0.005$ & $0.566 \pm 0.011$ \\
RAFT flow, $13$ statistics   & $0.868 \pm 0.004$ & $0.898 \pm 0.004$ & $0.851 \pm 0.004$ & $0.654 \pm 0.025$ \\
D3 features $+$ L2-LR        & $0.591 \pm 0.014$ & $0.665 \pm 0.006$ & $0.553 \pm 0.006$ & $0.552 \pm 0.004$ \\
CLIP ViT-B/32, PCA-$13$      & $0.869 \pm 0.005$ & $0.945 \pm 0.001$ & $0.852 \pm 0.004$ & $0.766 \pm 0.009$ \\
\bottomrule
\end{tabular}}
\end{table}

\begin{table}[htb]
\centering
\footnotesize
\setlength{\tabcolsep}{4pt}
\caption{\textbf{Seed spread and fold-level bootstrap $95\%$ intervals for rows of \cref{tab:identity}.} \gvb cell, LOGO OOD averaged over the $7$ held-out generators. Mean $\pm$ std is across the five seeds $\{42,0,7,17,99\}$. The interval is a percentile bootstrap over the seven held-out-generator folds. The floor is the seed-averaged \rvr through the same readout as the score in the same row. WaveRep's checkpoint is fixed, so its spread comes from the splits alone.}
\label{tab:supp_headline_ci}
\fitwidth{%
\begin{tabular}{@{}lccc@{}}
\toprule
Row (LOGO OOD) & Mean $\pm$ std & $95\%$ CI & Floor (\rvr) \\
\midrule
Codec-motion probe ($13$-d) & $0.8277 \pm 0.0028$ & $[0.7642, 0.8931]$ & $0.6279$ \\
RAFT flow, $13$ statistics  & $0.8512 \pm 0.0040$ & $[0.8227, 0.8840]$ & $0.6538$ \\
ReStraV ($21$-d)            & $0.9305 \pm 0.0014$ & $[0.8624, 0.9984]$ & $0.5803$ \\
D3 features $+$ L2-LR       & $0.5533 \pm 0.0065$ & $[0.5202, 0.5627]$ & $0.5518$ \\
\midrule
WaveRep head                & $0.9955 \pm 0.0005$ & $[0.9897, 0.9992]$ & $0.5334$ \\
WaveRep backbone            & $0.9984 \pm 0.0004$ & $[0.9979, 0.9997]$ & $0.7867$ \\
\bottomrule
\end{tabular}}
\end{table}

\section{Cells and Pipelines}
\label{app:cells}

A cell is a fixed set of real and generated clips on which the controls of \cref{tab:controls} run (\cref{sec:gates}). This section fixes the cells of the paper and the pipeline each went through.

\begin{table}[htb]
  \centering
  \footnotesize
  \setlength{\tabcolsep}{5pt}
  \renewcommand{\arraystretch}{0.96}
  \caption{\textbf{The controls.} Each control is one operation on one channel of \cref{fig:model}, or on the estimator (the readout) and the population (which collections a cell holds), which the figure does not draw. \emph{Check} is the condition fixed before any result was read. \emph{Requires} is what a corpus must supply.}
  \label{tab:controls}
  \fitwidth{%
  \begin{tabular}{@{}>{\raggedright\arraybackslash}p{1.05in}
                     >{\raggedright\arraybackslash}p{0.72in}
                     >{\raggedright\arraybackslash}p{1.55in}
                     >{\raggedright\arraybackslash}p{1.95in}
                     >{\raggedright\arraybackslash}p{1.3in}@{}}
    \toprule
    Channel & Operation & Control & Check & Requires \\
    \midrule
    container, encoding & intervene & one re-encode (the recipe) for every clip & the container probe (frame-type counts, no pixels) scores at chance on the clips the detectors see & the raw clips \\
    container, length & intervene & exactly $K$ non-I-frames per clip & as above & clips with $\geq K$ non-I-frames \\
    \addlinespace[1pt]
    signal, motion, and appearance & measure & the identity floor, the real-vs-real AUC through the detector's own features & codec-motion floor below $0.90$ (R3); a score that clears its floor by less than $0.15$ fails the screen & two independently curated real sources \\
    signal, motion, and appearance & pair & pairs sharing content, a fake against its source clip or a real clip against its reconstruction (\cref{sec:pairing}) & paired AUC beside the unpaired one & paired clips or a generator's autoencoder \\
    \addlinespace[1pt]
    estimator & hold fixed & the shared readout re-fitted for every compared feature set & every compared score uses the same readout & features or code \\
    estimator & quantify & seeds and fold-level intervals & intervals reported and gating every claim & folds with headroom \\
    \addlinespace[1pt]
    population & replicate & a second cell with its own real sources and generators & controls re-run per cell, transfer reported with intervals & an independent cell \\
    \bottomrule
  \end{tabular}}
\end{table}

\para{Pipelines.} Each feature set reads a cell through one of three pipelines, the recipe, the native pipeline, or the clips as shipped, and every table names which. The recipe is one canonical H.264 configuration, libx264~\citep{x264} at CRF $23$, GOP $12$, main profile (the common H.264 feature set), $8$\,fps, rescaled to $512{\times}512$ with FFmpeg's default bicubic filter, FFmpeg $7.1.1$~\citep{ffmpeg}. The fixed-pattern variant in the last column of \cref{tab:twobytwo} also switches off the encoder's shot-change I-frames and adaptive B-frame placement. It fixes two B-frames after every I- or P-frame and a GOP of exactly $12$, and the exact encoder options ship with the release. It then keeps only clips with a tail buffer of at least three frames past the window. On \aigvd, on the third cell, and on the census samples, the container and codec-motion probes read clips after the recipe. On the \gvb cell the codec-motion features come from the native pipeline, a $512{\times}512$ H.264 re-encode at the native frame rate with one I-frame per clip and at most $64$ frames. The shortest generators therefore contribute only their native $13$ to $16$ frames (MuseV $13$, Stable Video Diffusion $14$, ModelScope and VideoCrafter2 $16$). The deep detectors (CLIP, D3, WaveRep, ReStraV) read clips as their benchmark ships them, in their native codecs at $8$ to $30$\,fps, which is how the D3 and ReStraV papers score them. The re-extraction check of \cref{app:reprocess} confirms it, since a detector fitted on \aigvd flags about $5\%$ of that benchmark's real clips as shipped and far more of the same clips after the recipe. CLIP, WaveRep, and ReStraV are also re-extracted with every clip of both cells through the recipe (\cref{app:recipe}), and the census re-extracts the \gvb codec-motion features under the recipe (\cref{sec:census}).

The $K{=}12$ length filter applies to the rows that read motion over $K$ frames, the codec-motion probe and RAFT. The frame-based rows (D3, CLIP, WaveRep, and ReStraV) pool over decoded frames. The codec, frame rate, and frame size act on every frame they read, and the rows re-fitted under the recipe test how much that matters. Within a cell every row sees the same clips and differs only in what it computes. A per-method extraction audit on the \gvb cell records zero failed clips for FVMD, RAFT, or the codec-motion probe.

\para{\gvb.} The full set is the union of the benchmark's two releases, which it calls Pair\,1 and Pair\,2, $116{,}066$ clips with $26{,}956$ real clips from Vript and HD-VG-130M. The \gvb cell holds $27{,}000$ clips, $3{,}000$ per generator for CogVideo, Mora, ModelScope, MuseV, Pika, Stable Video Diffusion, and VideoCrafter2, and $3{,}000$ each from Vript and HD-VG-130M. Text2Video-Zero is excluded by the filter (\cref{app:gates}).

\para{\aigvd.} The cell is the Open-Source text-to-video split of \aigvd~\citep{aigvdbench2024}, which ships seven generators, AccVideo, AnimateDiff, CogVideoX-1.5, HunyuanVideo, Open-Sora, Pyramid-Flow, and Wan2.1, paired with $1{,}396$ real clips from one source, OpenVid-HD. The probes read every clip after the recipe, and the deep detectors read it as shipped, with the same extractors as on \gvb. Every detector is fitted and evaluated entirely within the cell, training on \aigvd splits and leaving one \aigvd generator out at a time. No \gvb weights are reused. At $K{=}12$ after the $8$ fps resample, AccVideo's clips are too short and HunyuanVideo keeps only $10$ clips, so both are dropped. Five generators remain, with $1{,}396$ real and $888$ generated clips. Relaxing the filter to $K{=}8$ keeps both, which gives the extended seven-generator cell of \cref{tab:supp_aigvd_cis}. The identity floor is undefined because the cell ships one real source.

\para{The third cell.} The third cell was built to a protocol frozen before any detector ran, and it fails gate R3. The protocol was frozen on 10 August 2026, and the dated protocol file ships with the release. Generated clips come from \genvideo-Val~\citep{chen2024demamba}. The two real pools are MSR-VTT, shipped inside \genvideo-Val, and a Kinetics-400 sample from the official mirror, curated independently of each other. Sampling is $500$ clips per source, uniformly at random without replacement at seed $42$ from the filename-sorted listing, with no selection on content, quality, or length and no inspection of detector scores at any point. After the recipe and the $K{=}12$ filter, $4{,}906$ clips survive. The $3{,}911$ generated clips come from eight generators: Gen2 ($500$), Show-1 ($500$), ModelScope ($497$), MorphStudio ($497$), VideoCrafter ($488$), LaVie ($486$), MoonValley ($479$), and \genvideo's WildScrape source ($464$). The $995$ real clips are $497$ from MSR-VTT and $498$ from Kinetics-400. Two sources were excluded by rule before any score was computed. HotShot ships one-second clips and none survives $K{=}12$. Sora has $56$ clips, below the $100$-clip minimum of gate R2. Through the codec-motion probe under the shared readout, over five seeds, the cell gives LOGO ID $0.830$, LOGO OOD $0.806 \pm 0.011$, a floor of $0.976$, and a margin of $-0.171$. R1 and R2 pass and R3 fails (\cref{sec:census}). MSR-VTT and Kinetics-400 are near-perfectly separable by codec motion alone, so the readout finds the corpus boundary more easily than the label, and nothing measured on the cell is interpretable as detection. We publish no ranking on it.

\para{How the two benchmarks differ.} \gvb and \aigvd differ in their generators, in their pipelines, and in size, since \aigvd's LOGO folds hold at most $1{,}600$ training clips. Pipeline and generators alone move the codec-motion probe by about $0.1$ on \gvb itself, from $0.832$ under the native pipeline with seven generators to $0.736$ under the recipe with five (\cref{fig:transfer}).

\begin{table}[htb]
\centering
\footnotesize
\setlength{\tabcolsep}{3.5pt}
\caption{\textbf{The container probe on every corpus and processing condition.} LOGO AUC with fold-bootstrap $95\%$ intervals for the three-feature probe (I-, P-, and B-frame counts) and the one-feature probe (total frames). Fixed: the fixed frame-type pattern of \cref{sec:leakage}. Tail buffer: clips of at least $16$ re-encoded frames.}
\label{tab:supp_container_grid}
\fitwidth{%
\begin{tabular}{@{}llcccccc@{}}
\toprule
Corpus & Probe & Shipped & \cistack{Length}{filter} & Re-encode & Both & \cistack{Both,}{fixed} & \cistack{Both, fixed,}{tail buffer} \\
\midrule
\gvb & I, P, B counts & \cistack{$0.999$}{[1.00, 1.00]} & \cistack{$0.578$}{[0.37, 0.71]} & \cistack{$0.975$}{[0.95, 0.99]} & \cistack{$0.481$}{[0.41, 0.55]} & \cistack{$0.500$}{[0.50, 0.50]} & \cistack{$0.500$}{[0.50, 0.50]} \\
\gvb & total frames & \cistack{$0.998$}{[0.99, 1.00]} & \cistack{$0.420$}{[0.29, 0.49]} & \cistack{$0.986$}{[0.97, 1.00]} & \cistack{$0.528$}{[0.52, 0.53]} & \cistack{$0.500$}{[0.50, 0.50]} & \cistack{$0.500$}{[0.50, 0.50]} \\
\aigvd & I, P, B counts & \cistack{$0.997$}{[0.99, 1.00]} & \cistack{$0.945$}{[0.93, 0.96]} & \cistack{$0.929$}{[0.83, 1.00]} & \cistack{$0.633$}{[0.48, 0.79]} & \cistack{$0.700$}{[0.50, 0.90]} & \cistack{$0.500$}{[0.50, 0.50]} \\
\aigvd & total frames & \cistack{$0.989$}{[0.97, 1.00]} & \cistack{$0.496$}{[0.49, 0.50]} & \cistack{$0.927$}{[0.83, 1.00]} & \cistack{$0.496$}{[0.49, 0.50]} & \cistack{$0.500$}{[0.50, 0.50]} & \cistack{$0.500$}{[0.50, 0.50]} \\
\genvideo{} + K400 & I, P, B counts & \cistack{$0.839$}{[0.66, 0.95]} & \cistack{$0.704$}{[0.65, 0.75]} & \cistack{$0.952$}{[0.88, 1.00]} & \cistack{$0.613$}{[0.57, 0.67]} & \cistack{$0.549$}{[0.50, 0.65]} & \cistack{$0.500$}{[0.50, 0.50]} \\
\genvideo{} + K400 & total frames & \cistack{$0.777$}{[0.60, 0.93]} & \cistack{$0.637$}{[0.62, 0.65]} & \cistack{$0.948$}{[0.86, 1.00]} & \cistack{$0.575$}{[0.56, 0.58]} & \cistack{$0.500$}{[0.50, 0.50]} & \cistack{$0.500$}{[0.50, 0.50]} \\
\dvf & I, P, B counts & \cistack{$0.893$}{[0.81, 0.96]} & \cistack{$0.785$}{[0.62, 0.91]} & \cistack{$0.800$}{[0.71, 0.86]} & \cistack{$0.625$}{[0.54, 0.70]} & \cistack{$0.505$}{[0.505, 0.505]} & \cistack{$0.500$}{[0.50, 0.50]} \\
\dvf & total frames & \cistack{$0.858$}{[0.76, 0.94]} & \cistack{$0.756$}{[0.59, 0.86]} & \cistack{$0.786$}{[0.71, 0.85]} & \cistack{$0.548$}{[0.54, 0.55]} & \cistack{$0.500$}{[0.50, 0.50]} & \cistack{$0.500$}{[0.50, 0.50]} \\
\bottomrule
\end{tabular}}
\end{table}

\section{The Controls}
\label{app:protocol}

Each control is one operation on one channel of \cref{fig:model}, or on the estimator that turns features into a score (\cref{tab:controls}). For each, the table gives the check fixed before any result was read and what a corpus must supply for the control to run. The VidAudit toolkit runs these controls on any corpus, from the recipe and the length filter to the shared readout with floors, margins, seeds, and fold intervals. Its input is a manifest, a CSV file of clip paths, labels, and sources. Pairing runs on the natural pairs of two benchmarks (\cref{sec:pairing}).

\section{Container Probe Details}
\label{app:container}

The \aigvd row of \cref{tab:twobytwo} depends on the sample and the frame rate, and the full census grid with intervals is in \cref{tab:supp_container_grid}. We re-ran the probe on \aigvd at the native frame rate, over the six generators and the real pool present in both its native and re-encoded copies. There the three-feature probe gives $0.9988$ with neither control, $0.9473$ with the filter alone, $0.9890$ with the re-encode alone, and $0.4918$ with both. The one-feature probe on the total count falls to $0.5024$ under the filter. The census differs in sample, roster, and frame rate, and leaves $0.633$ under both controls.

The grid counts frame types at the codec level, which differs slightly from the length filter's count. A frame counts as non-I when its bitstream picture type is P or B, whereas the length filter counts frames that carry non-zero motion-vector magnitude. The two agree except on P- and B-frames that happen to carry no motion, which is why the grid applies an explicit truncation of its own.

On \gvb nearly all of the leak is clip length, because the benchmark's GOP structure is already uniform (\cref{tab:supp_gop}). Real clips carry hundreds of P- and B-frames and generated clips a few dozen at most, so the raw totals separate the classes by themselves. The within-clip ratios of I-, P-, and B-frames sit in narrow bands across every source as shipped, and the filter alone leaves $0.578$. On \aigvd as shipped the GOP structure is not uniform, which is why that benchmark needs both controls.

\begin{table}[htb]
\centering
\footnotesize
\setlength{\tabcolsep}{3pt}
\caption{\textbf{Per-source frame types on the \gvb cell as shipped.} Counts are mean frames per clip; $\widetilde{r_X}$ is the median over clips of $n_X/n_\text{total}$. The ratios are near-uniform across sources. The raw totals separate real from generated because generated clips are far shorter.}
\label{tab:supp_gop}
\fitwidth{%
\begin{tabular}{@{}lcccccccc@{}}
\toprule
Source & Real & $n$ & $\overline{n_I}$ & $\overline{n_P}$ & $\overline{n_B}$ & $\widetilde{r_I}$ & $\widetilde{r_P}$ & $\widetilde{r_B}$ \\
\midrule
HD-VG-130M & yes & 3000 & 8.35 & 269.3 & 490.9 & 0.009 & 0.327 & 0.660 \\
Vript      & yes & 3000 & 2.85 & 106.3 & 213.1 & 0.008 & 0.309 & 0.680 \\
\midrule
CogVideo   & no & 3000 & 1.34 & 10.3 & 21.4 & 0.030 & 0.242 & 0.727 \\
Mora       & no & 3000 & 1.08 &  6.9 & 16.1 & 0.042 & 0.250 & 0.708 \\
ModelScope & no & 3000 & 1.32 &  4.9 &  9.8 & 0.062 & 0.250 & 0.688 \\
MuseV      & no & 3000 & 1.00 &  3.4 &  8.6 & 0.077 & 0.231 & 0.692 \\
Pika       & no & 3000 & 1.12 & 19.4 & 39.0 & 0.014 & 0.278 & 0.694 \\
Stable Video Diffusion & no & 3000 & 1.11 &  4.4 &  8.5 & 0.071 & 0.286 & 0.643 \\
VideoCrafter2 & no & 3000 & 1.16 &  4.2 & 10.6 & 0.062 & 0.250 & 0.688 \\
\bottomrule
\end{tabular}}
\end{table}

\para{The full grid.} \cref{tab:supp_container_grid} extends \cref{tab:twobytwo} to both probes and every processing condition. The length filter keeps the prefix that ends at the twelfth non-I-frame. With the fixed frame-type pattern (no shot-change I-frames, two B-frames after every I- or P-frame, and a GOP of $12$) and with a tail buffer of three frames past that prefix, the three-feature probe is at chance on all four corpora. The buffer is needed because the encoder rewrites the last B-frames of a stream as P-frames. The tail buffer is a selection. Of the clips that keep $K$ non-I-frames after the fixed re-encode, it keeps 2,997 of 3,499 on \gvb, 2,033 of 3,033 on \aigvd, 4,576 of 5,031 on \genvideo{} + K400, and 3,764 of 3,771 on \dvf.

\section{Census Details}
\label{app:censusdetail}

\para{Sampling.} The census reads every corpus through one manifest, released with the code, with one row per source. For each source it draws $500$ clips uniformly without replacement from the filename-sorted listing at seed $42$, or every clip when a source ships fewer. \gvb sources are restricted to the clips of the \gvb cell. Each clip is probed as shipped and after the recipe, and its frame types are read from the frame headers. \cref{tab:supp_census_sources,tab:supp_census_sources_b} list the sources, one row each, with source names as each benchmark's release gives them. For every source they give the clips decoded out of the $500$ sampled, the majority native codec, the median frame rate, the most common resolution, and the median native length in frames. The last two columns give the share of clips that keep $K{=}12$ non-I-frames as shipped and after the recipe. These shares count P- and B-frames by picture type, while the cells' length filter counts frames with non-zero motion vectors (\cref{app:container}), so a cell can keep fewer clips than the share implies.

\begin{table}[htb]
\centering
\scriptsize
\setlength{\tabcolsep}{4pt}
\caption{\textbf{Census sources of \gvb and \aigvd.} One row per source; columns as described in \cref{app:censusdetail}.}
\label{tab:supp_census_sources}
\fitwidth{%
\begin{tabular}{@{}lllrlrlrrr@{}}
\toprule
Corpus & Source & Side & Clips & Codec & fps & Resolution & Length & $K$ shipped (\%) & $K$ re-encoded (\%) \\
\midrule
\gvb & HD-VG-130M & real & 500 & H.264 & 30 & 1280$\times$720 & 370 & 100 & 100 \\
 & Vript & real & 500 & H.264 & 30 & 1280$\times$720 & 184 & 100 & 100 \\
\addlinespace[3pt]
 & CogVideo & gen. & 500 & H.264 & 8 & 480$\times$480 & 33 & 100 & 100 \\
 & Mora & gen. & 500 & H.264 & 10 & 1024$\times$576 & 24 & 100 & 100 \\
 & ModelScope & gen. & 500 & MPEG-4 & 8 & 256$\times$256 & 16 & 100 & 100 \\
 & MuseV & gen. & 500 & H.264 & 12 & 1216$\times$640 & 13 & 99 & 0 \\
 & Pika & gen. & 500 & H.264 & 24 & 1088$\times$640 & 72 & 100 & 100 \\
 & SVD & gen. & 500 & H.264 & 10 & 1280$\times$704 & 14 & 100 & 0 \\
 & Text2Video-Zero & gen. & 500 & H.264 & 4 & 512$\times$512 & 8 & 0 & 6 \\
 & VideoCrafter2 & gen. & 500 & H.264 & 10 & 512$\times$320 & 16 & 100 & 99 \\
\midrule
\aigvd & real pool & real & 500 & H.264 & 30 & 1920$\times$1080 & 158 & 100 & 100 \\
\addlinespace[3pt]
 & AccVideo & gen. & 500 & H.264 & 24 & 960$\times$544 & 33 & 100 & 0 \\
 & AnimateDiff & gen. & 500 & H.264 & 10 & 512$\times$512 & 16 & 100 & 100 \\
 & CogVideoX-1.5 & gen. & 500 & H.264 & 16 & 720$\times$480 & 33 & 100 & 100 \\
 & HunyuanVideo & gen. & 500 & H.264 & 24 & 960$\times$544 & 33 & 100 & 7 \\
 & Open-Sora & gen. & 500 & H.264 & 24 & 336$\times$192 & 37 & 100 & 100 \\
 & Pyramid-Flow & gen. & 500 & H.264 & 24 & 640$\times$384 & 89 & 100 & 100 \\
 & Wan2.1 & gen. & 500 & H.264 & 16 & 832$\times$480 & 81 & 100 & 100 \\
\bottomrule
\end{tabular}}
\end{table}

\para{\dvf's named sources.} \dvf names two real sources, YouTube-8M~\citep{abuelhaija2016youtube8m} and InternVid-10M~\citep{wang2024internvid}. InternVid is itself drawn from YouTube, and every real clip in the release carries a YouTube id, so no label says which of the two sources a clip came from. \cref{tab:supp_census_sources_b} lists \dvf's YouTube pool and its generators.

\para{Further census readings.} A six-way classifier on the thirteen codec-motion features names the pool of a real clip $40\%$ of the time, against a chance rate of $17\%$. The motion and appearance channels disagree where they should. \dvf's pool and Kinetics-400 are close in motion ($0.55$) but far apart in appearance ($0.88$), as two collections from one platform with different content would be.

\begin{table}[htb]
\centering
\scriptsize
\setlength{\tabcolsep}{4pt}
\caption{\textbf{Census sources of \genvideo{} + K400 and \dvf.} Columns as in \cref{tab:supp_census_sources}.}
\label{tab:supp_census_sources_b}
\fitwidth{%
\begin{tabular}{@{}lllrlrlrrr@{}}
\toprule
Corpus & Source & Side & Clips & Codec & fps & Resolution & Length & $K$ shipped (\%) & $K$ re-encoded (\%) \\
\midrule
\genvideo{} + K400 & MSR-VTT & real & 500 & H.264 & 3 & 298$\times$224 & 41 & 100 & 100 \\
 & Kinetics-400 & real & 500 & H.264 & 30 & 1280$\times$720 & 300 & 100 & 100 \\
\addlinespace[3pt]
 & VideoCrafter & gen. & 500 & H.264 & 10 & 1024$\times$640 & 16 & 100 & 100 \\
 & Gen2 & gen. & 500 & H.264 & 24 & 1408$\times$768 & 96 & 100 & 100 \\
 & HotShot & gen. & 500 & H.264 & 8 & 672$\times$384 & 8 & 0 & 0 \\
 & LaVie & gen. & 500 & H.264 & 8 & 512$\times$320 & 16 & 100 & 100 \\
 & ModelScope & gen. & 500 & H.264 & 8 & 1280$\times$720 & 32 & 100 & 100 \\
 & MoonValley & gen. & 500 & H.264 & 50 & 1184$\times$672 & 91 & 100 & 100 \\
 & MorphStudio & gen. & 500 & H.264 & 8 & 1024$\times$576 & 16 & 100 & 100 \\
 & Show-1 & gen. & 500 & H.264 & 8 & 576$\times$320 & 29 & 100 & 100 \\
 & Sora & gen. & 56 & H.264 & 30 & 1920$\times$1080 & 450 & 100 & 100 \\
 & WildScrape & gen. & 500 & H.264 & 8 & 512$\times$512 & 25 & 96 & 94 \\
\midrule
\dvf & YouTube pool & real & 500 & MPEG-4 & 30 & 1280$\times$720 & 484 & 100 & 93 \\
\addlinespace[3pt]
 & CogVideoX & gen. & 500 & H.264 & 16 & 1360$\times$768 & 81 & 100 & 100 \\
 & HunyuanVideo & gen. & 500 & H.264 & 30 & 1280$\times$720 & 129 & 100 & 100 \\
 & LDM & gen. & 15 & H.264 & 30 & 1080$\times$1920 & 358 & 100 & 100 \\
 & Open-Sora & gen. & 500 & H.264 & 8 & 512$\times$512 & 16 & 100 & 100 \\
 & Pika & gen. & 103 & H.264 & 24 & 1280$\times$720 & 72 & 100 & 100 \\
 & Sora & gen. & 87 & H.264 & 30 & 1024$\times$576 & 300 & 100 & 95 \\
 & StableVideo & gen. & 104 & H.264 & 24 & 1024$\times$576 & 97 & 100 & 99 \\
 & SVD & gen. & 500 & MPEG-4 & 7 & 1024$\times$576 & 14 & 100 & 100 \\
 & VideoCrafter & gen. & 500 & H.264 & 8 & 1024$\times$576 & 16 & 100 & 100 \\
 & ZeroScope & gen. & 500 & MPEG-4 & 8 & 1024$\times$576 & 36 & 100 & 100 \\
\bottomrule
\end{tabular}}
\end{table}

\section{Fold-Level Intervals for the Replication Column}
\label{app:foldci}

Every entry of the \aigvd column of \cref{tab:identity} carries a fold-level interval, because the \aigvd folds are small (\cref{tab:supp_aigvd_cis}). Each holds between $28$ and $40$ held-out clips, so a fold mean can hide a wide spread. Each detector's per-generator LOGO scores come from the readout of \cref{app:ci}, and folds are resampled with the same percentile bootstrap used for the \gvb rows ($1{,}000$ resamples, seed $42$).

\begin{table}[htb]
\centering
\footnotesize
\setlength{\tabcolsep}{8pt}
\caption{\textbf{Fold-level bootstrap intervals on the \aigvd column of \cref{tab:identity}.} Percentile bootstrap over held-out generators, $1{,}000$ resamples, seed $42$, shared readout. \emph{Range} is the observed spread across folds. The last column counts folds at or below chance. The last row is the codec-motion probe on the extended seven-generator cell at $K{=}8$ (\cref{app:cells}).}
\label{tab:supp_aigvd_cis}
{%
\begin{tabular}{@{}lcccc@{}}
\toprule
Detector & Mean & $95\%$ CI & Range & $\leq 0.5$ \\
\midrule
WaveRep head                  & $0.981$ & $[0.959,0.998]$ & $0.940$--$1.000$ & 0 \\
CLIP ViT-B/32, PCA-$13$       & $0.750$ & $[0.701,0.828]$ & $0.691$--$0.902$ & 0 \\
D3 ($768$-d features)         & $0.722$ & $[0.686,0.754]$ & $0.652$--$0.765$ & 0 \\
ReStraV ($21$-d)              & $0.673$ & $[0.422,0.894]$ & $0.224$--$0.941$ & 1 \\
RAFT flow, $13$ statistics    & $0.648$ & $[0.622,0.679]$ & $0.612$--$0.709$ & 0 \\
FVMD (full $1024$-d)          & $0.606$ & $[0.566,0.659]$ & $0.549$--$0.709$ & 0 \\
Codec-motion probe ($13$-d)   & $0.602$ & $[0.528,0.676]$ & $0.492$--$0.702$ & 1 \\
\midrule
Codec-motion probe, $7$-generator cell ($K{=}8$) & $0.667$ & $[0.588,0.758]$ & $0.495$--$0.878$ & 1 \\
\bottomrule
\end{tabular}}
\end{table}

The intervals separate WaveRep from every other detector and leave ReStraV indistinguishable from chance. WaveRep's is tight and clears every other detector's upper bound, so its replication on \aigvd is the one the folds support outright. ReStraV's spans chance, with one fold at $0.224$ where the detector is anti-predictive, so its \aigvd score is better described as indistinguishable from chance than by its point estimate. Because WaveRep's lower bound of $0.959$ lies above ReStraV's upper bound of $0.894$, the separation between the two rests on non-overlapping intervals. The codec-motion probe's lower bound also sits near chance, and the appearance probe falls by $0.10$ with an upper bound of $0.828$, below its \gvb score of $0.852$. The CLIP row uses the PCA-$13$ readout of \cref{tab:identity}, the readout every CLIP score in this paper uses.

ReStraV's anti-predictive fold is Open-Sora. Re-scored per generator through the same path at five seeds, ReStraV reaches $0.234$ on held-out Open-Sora (the $0.224$ above is seed $42$) and $0.484$ on Pyramid-Flow, against $0.78$ to $0.94$ on the other three generators. Over five seeds, two of the five \aigvd generators score at or below chance, which is why ReStraV's interval spans chance. At seed $42$ only Open-Sora does, so \cref{tab:supp_aigvd_cis} counts one such fold.

\section{The Margin Cut and the Length Filter}
\label{app:gates}

\para{Margin cut.} A score on \gvb passes the screen when it clears its own floor by $0.15$ or more. We swept this cut over $[0.10, 0.20]$ in steps of $0.01$ and re-derived each row's label from the scores of \cref{tab:identity}. Seven of the eight upper rows of \cref{tab:identity} keep their label at every cut, because their margins lie outside the band: CLIP ($+0.086$) and D3's features ($+0.011$) below it, and the other five at $+0.22$ or more above it. The codec-motion probe is the one row that moves. Its margin of $+0.189$ sits one thousandth below a cut of $0.19$, so it clears the cut up to $0.18$ and falls short from $0.19$. We report the move and keep the cut. The R3 threshold of $0.90$ gives the same gate outcomes anywhere between the \gvb floor of $0.643$ and the third cell's $0.976$.

\para{Length filter.} The length filter keeps clips with at least $K$ non-I-frames and reads the first $K$, so $K$ decides which generators a cell admits and how much temporal support each feature has. On a fixed $9{,}000$-clip \gvb subset the codec-motion probe gives LOGO OOD $0.771$ at $K{=}8$, $0.825$ at $K{=}12$, and $0.831$ at $K{=}16$. Because $K{=}16$ discards $45\%$ of the clips, $K{=}12$ is the operating point. Text2Video-Zero fails the filter at every $K \geq 8$ (none of its $3{,}000$ clips survive $K{=}8$), so its exclusion is structural. \cref{tab:supp_k_paired} compares $K{=}12$ with $K{=}8$ on identical clips. Both are recomputed from the same stored motion-vector arrays, so $K$ is the only variable. The recomputed $K{=}12$ features match those of the main text to $\max|\Delta| = 6.9\times10^{-7}$ over all $27{,}000$ clips and thirteen features.

\begin{table}[htb]
\centering
\footnotesize
\setlength{\tabcolsep}{4pt}
\caption{\textbf{Paired $K$ sensitivity on identical clips.} Both rows are recomputed from the same stored motion-vector arrays, so $K$ is the only variable. The AUC moves by about three points and the margin by one. Two clips yield no features at $K{=}8$ and are dropped.}
\label{tab:supp_k_paired}
\begin{tabular}{@{}lccccc@{}}
\toprule
$K$ & Clips & LOGO ID & LOGO OOD & Floor (\rvr) & Margin \\
\midrule
$12$ & $27{,}000$ & $0.9088$ & $0.8324$ & $0.6425$ & $+0.1898$ \\
$8$  & $26{,}998$ & $0.8908$ & $0.8012$ & $0.6215$ & $+0.1796$ \\
\midrule
$\Delta$ & $-2$ & $-0.0180$ & $-0.0312$ & $-0.0210$ & $\mathbf{-0.0102}$ \\
\bottomrule
\end{tabular}
\end{table}

Lowering $K$ to $8$ moves the margin by one point, because the floor falls with the score (\cref{tab:supp_k_paired}). LOGO OOD falls by $3.1$ AUC points and the floor by $2.1$, so the margin, the quantity a gate outcome is computed from, moves by $1.0$. No row of \cref{tab:identity} changes its gate outcome under a $0.01$ margin shift (the sweep above spans ten times that).

\section{Survey of Controls in 20 Detection Papers}
\label{app:survey}

We surveyed $20$ papers on detecting or evaluating AI-generated video from $2024$ to $2026$ for six checks drawn from the controls of \cref{tab:controls}. The sample is not systematic. It holds every method and benchmark used in this work and a broader set of recent detectors and benchmarks for fully synthesized video. Face-manipulation deepfake detectors, which target face swapping and reenactment, are out of scope. Each paper was coded from its full text and appendices in three independent passes, each quoting the passage behind its code. Each table entry takes the majority code, and $105$ of the $120$ entries were unanimous. \cref{tab:supp_survey} gives the result, and the quoted evidence ships with the release.

\begin{table}[htb]
\centering
\scriptsize
\setlength{\tabcolsep}{5pt}
\caption{\textbf{Methodological controls in $20$ recent papers on AI-generated video detection.} Re-encode: inputs through one fixed codec pipeline. Container audit: a trivial container probe on the clips the detectors see (the check of the re-encode and length filter in \cref{tab:controls}). Floor: a real-vs-real floor between two real pools. Matched readout: compared methods re-trained under one protocol. Seeds/CI: multi-seed stability or bootstrap intervals. Cross-corpus: evaluation on a benchmark with independently curated real clips and generators. No paper applies all six, and the container audit, the floor, and seeds or intervals are present in full in at most two papers each. \protect\hbfull\ present, \protect\hbhalf\ partial (for example re-training only some baselines, or a same-dataset sanity check for the floor), \protect\hbnone\ absent.}
\label{tab:supp_survey}
\fitwidth{%
\begin{tabular}{@{}lcccccc@{}}
\toprule
Paper & Re-encode & Container audit & Floor & Matched readout & Seeds/CI & Cross-corpus \\
\midrule
D3~\citep{ma2024d3}                & \hbnone & \hbnone & \hbnone & \hbhalf & \hbnone & \hbfull \\
ReStraV~\citep{restrav2024}         & \hbnone & \hbnone & \hbnone & \hbnone & \hbnone & \hbfull \\
DeCoF~\citep{maaz2024decof}         & \hbnone & \hbhalf & \hbnone & \hbhalf & \hbnone & \hbhalf \\
ATSS~\citep{xu2024atss}             & \hbnone & \hbnone & \hbnone & \hbhalf & \hbnone & \hbhalf \\
CMTA~\citep{zhao2024cmta}           & \hbnone & \hbnone & \hbnone & \hbhalf & \hbnone & \hbfull \\
VidGuard-R1~\citep{park2025vidguard}& \hbnone & \hbnone & \hbnone & \hbhalf & \hbnone & \hbfull \\
FVMD~\citep{liu2024fvmd}            & \hbnone & \hbnone & \hbhalf\textsuperscript{$\dagger$} & \hbnone & \hbnone & \hbnone \\
\aigvd~\citep{aigvdbench2024}       & \hbfull & \hbhalf & \hbnone & \hbhalf & \hbnone & \hbhalf \\
\gvb~\citep{ni2025genvidbenchchallengingbenchmarkdetecting} & \hbnone & \hbnone & \hbnone & \hbhalf & \hbnone & \hbhalf \\
AIGVDet~\citep{bai2024aigvdet}      & \hbnone & \hbnone & \hbnone & \hbhalf & \hbnone & \hbhalf \\
DeMamba~\citep{chen2024demamba}     & \hbnone & \hbnone & \hbnone & \hbfull & \hbnone & \hbhalf \\
LGTD~\citep{he2024lgtd}             & \hbnone & \hbnone & \hbnone & \hbhalf & \hbnone & \hbhalf \\
Native-Scale Forgery~\citep{li2026nativescale}  & \hbnone & \hbnone & \hbnone & \hbhalf & \hbnone & \hbfull \\
VideoVeritas~\citep{tan2026videoveritas}        & \hbnone & \hbnone & \hbnone & \hbhalf & \hbnone & \hbfull \\
Learned 3D evaluation~\citep{chang2024whatmatters} & \hbnone & \hbnone & \hbnone & \hbnone & \hbnone & \hbfull \\
NSG-VD~\citep{zhang2025nsgvd}       & \hbnone & \hbnone & \hbnone & \hbfull & \hbfull & \hbfull \\
WaveRep~\citep{corvi2025waverep}    & \hbfull & \hbhalf & \hbnone & \hbfull & \hbnone & \hbfull \\
UNITE~\citep{kundu2025unite}        & \hbnone & \hbnone & \hbnone & \hbnone & \hbnone & \hbfull \\
Deepfake-Eval-2024~\citep{chandra2025deepfakeeval} & \hbnone & \hbnone & \hbnone & \hbhalf & \hbnone & \hbfull \\
RobustSora~\citep{wang2025robustsora}            & \hbnone & \hbfull & \hbnone & \hbfull & \hbfull & \hbhalf \\
\bottomrule
\end{tabular}}\par\smallskip
{\scriptsize \textsuperscript{$\dagger$}\,FVMD reports a same-dataset distributional convergence check (the BAIR robot-pushing videos against themselves, FVMD $\approx 0$), a sanity check with no real-vs-real classifier.\par}
\end{table}

\para{Patterns.} Cross-corpus evaluation is common in these papers, and the controls that probe identity are nearly absent. Nearly every paper tests on held-out generators, and roughly half also test on an independently curated benchmark. Re-training compared methods under one protocol is the next most common and the most variable. Four papers re-train all compared methods under one protocol (DeMamba~\citep{chen2024demamba}, NSG-VD~\citep{zhang2025nsgvd}, WaveRep~\citep{corvi2025waverep}, and RobustSora~\citep{wang2025robustsora}), most re-train only some baselines while citing the rest, and a few cite all published numbers. A canonical re-encode is rare. Only \aigvd~\citep{aigvdbench2024} and WaveRep~\citep{corvi2025waverep} re-encode every input through one fixed codec pipeline before feature extraction. The remaining three controls are nearly absent. A container audit against a trivial probe appears in full only in RobustSora~\citep{wang2025robustsora}, with partial shortcut analyses in a few others. The real-vs-real floor appears in no paper in full form, the closest being FVMD's same-dataset convergence check~\citep{liu2024fvmd}. Multi-seed or bootstrap intervals appear in two papers. RobustSora~\citep{wang2025robustsora} reports a $3$-seed mean $\pm$ std and a $10{,}000$-sample paired bootstrap on its own benchmark. NSG-VD~\citep{zhang2025nsgvd} reports, in its Appendix D.1, standard deviations of four metrics across ten datasets at three seeds. No paper applies all six. The closest, WaveRep and RobustSora, each apply three checks fully and one partially, and neither reports a real-vs-real floor.

\section{Adversaries and Pairing}
\label{app:adaptive}
\label{sec:adaptive}
\label{sec:pairing}

Two adversaries who know the controls match the generated side to the real side and ask which detectors still separate them. The re-encode and the length filter are already applied to the probes' feature tables, so we test two further matchings on each cell. Each matching subsamples both sides without replacement to one common histogram, so that no clip is duplicated across later splits. The first matches mean motion-vector magnitude, the bias that \citet{michels2026motionbias} report for motion-based detectors. The second is sharper. It matches the generated side on the codec-motion probe's own out-of-fold score. The probe then separates the two sides at chance by construction, and the question becomes which other detectors still do. Every detector with features on the cell is then re-fitted under the shared readout on the matched clips, five seeds each (\cref{tab:adaptive}).

\begin{table}[htb]
  \centering
  \footnotesize
  \setlength{\tabcolsep}{3.5pt}
  \renewcommand{\arraystretch}{1.02}
  \caption{\textbf{Two adversaries who know the controls.} LOGO OOD, mean over five seeds, on each cell's clips (base), after matching mean motion magnitude (motion), and after matching the codec-motion probe's score (probe).}
  \label{tab:adaptive}
  \fitwidth{%
  \begin{tabular}{@{}lcccccc@{}}
    \toprule
    & \multicolumn{3}{c}{\gvb} & \multicolumn{3}{c}{\aigvd} \\
    \cmidrule(lr){2-4}\cmidrule(l){5-7}
    Detector & Base & Motion & Probe & Base & Motion & Probe \\
    \midrule
    WaveRep head & $0.996$ & $0.995$ & $0.995$ & $0.983$ & $0.983$ & $0.982$ \\
    ReStraV & $0.931$ & $0.933$ & $0.915$ & $0.672$ & $0.683$ & $0.687$ \\
    FVMD & $0.886$ & $0.875$ & $\mathbf{0.787}$ & -- & -- & -- \\
    RAFT flow statistics & $0.851$ & $0.849$ & $\mathbf{0.792}$ & -- & -- & -- \\
    CLIP, PCA-13 & $0.824$ & $0.829$ & $0.814$ & $0.732$ & $0.723$ & $0.745$ \\
    Codec-motion probe & $0.828$ & $0.827$ & -- & $0.597$ & $0.581$ & -- \\
    D3 features & $0.553$ & $0.547$ & $0.544$ & -- & -- & -- \\
    \bottomrule
  \end{tabular}}
\end{table}

Matching motion magnitude moves no detector, and matching the codec-motion probe's score lowers only FVMD and RAFT (\cref{tab:adaptive}). The base column is this experiment's own baseline on the same clips. CLIP's PCA is fitted without labels on the whole table here, so its baseline is transductive. The second matching equalizes the probe's own out-of-fold score, so the probe has no entry there, and FVMD, RAFT, and D3 were not part of the \aigvd run. Once the length filter is applied, mean motion magnitude alone separates real from generated at $0.48$ on \gvb and $0.49$ on \aigvd, so the bias Michels et al.\ report is absent here, and matching it moves no detector by more than $0.016$. Equalizing the generated side on the codec-motion probe's score lowers FVMD from $0.886$ to $0.787$ and RAFT from $0.851$ to $0.792$. These drops are about a third of FVMD's margin of $+0.306$ and a quarter of RAFT's $+0.228$. ReStraV, CLIP, D3, and WaveRep's head move by at most $0.016$ on either cell. FVMD and RAFT read the same motion information the probe does. The others read different signal, so an adversary who neutralizes one channel leaves the detectors that read another untouched, and the floor of each detector has to be measured through its own features.

\para{Pairing.} \aigvd makes every generated clip from an identified real clip and reuses each real id across its generators, and \dvf pairs fakes with real partners within each generator's folder. For each generator, half of its pairs are held out, and the readout is fitted on the remaining real clips and the other generators' fakes. The paired AUC is the share of held-out pairs in which the fake scores above its own source clip. \cref{tab:paired} gives it beside the unpaired AUC on the same clips. The largest drop for any generator is $0.031$, and CLIP's paired score is higher for every generator, so content variation between the two sides does not carry these scores. Identity enters between collections, where a paired test inside one benchmark cannot see it. The deep detectors read clips as shipped, with $150$ held-out pairs per generator over five seeds. The codec-motion probe reads the recipe, with $120$ to $150$ held-out pairs per generator at seed $42$, on the generators that keep $K$ frames.

\begin{table}[htb]
  \centering
  \footnotesize
  \setlength{\tabcolsep}{4pt}
  \renewcommand{\arraystretch}{1.02}
  \caption{\textbf{Pairing leaves the within-benchmark scores in place.} Paired AUC (each fake against its own source clip) and unpaired AUC on the same clips, means over generators, with the range of paired minus unpaired over generators.}
  \label{tab:paired}
  \fitwidth{%
  \begin{tabular}{@{}lcccccc@{}}
    \toprule
    & \multicolumn{3}{c}{\aigvd} & \multicolumn{3}{c}{\dvf} \\
    \cmidrule(lr){2-4}\cmidrule(l){5-7}
    Detector & Paired & Unpaired & $\Delta$ range & Paired & Unpaired & $\Delta$ range \\
    \midrule
    CLIP, PCA-13 & $0.746$ & $0.704$ & $+0.012$ to $+0.085$ & $0.911$ & $0.867$ & $+0.003$ to $+0.144$ \\
    ReStraV & $0.744$ & $0.747$ & $-0.029$ to $+0.019$ & $0.721$ & $0.729$ & $-0.030$ to $+0.006$ \\
    WaveRep head & $0.990$ & $0.987$ & $-0.003$ to $+0.017$ & $0.993$ & $0.992$ & $-0.001$ to $+0.001$ \\
    WaveRep backbone & $1.000$ & $1.000$ & $-0.001$ to $+0.001$ & $0.996$ & $0.995$ & $-0.002$ to $+0.005$ \\
    Codec-motion probe & $0.631$ & $0.640$ & $-0.023$ to $+0.006$ & $0.765$ & $0.774$ & $-0.031$ to $+0.024$ \\
    \midrule
    Generators & \multicolumn{3}{c}{7 (codec-motion probe: 5)} & \multicolumn{3}{c}{6} \\
    \bottomrule
  \end{tabular}}
\end{table}

\section{Re-processing Real Clips}
\label{app:reprocess}

Each processing step moves the false-positive rate of a different detector (\cref{tab:reprocess}, behind \cref{sec:reprocess}). Each feature set is fitted on its cell with every clip as shipped and thresholded at $5\%$ false positives on held-out real clips as shipped. The six real collections are then scored in four forms. The first is as shipped, the second a plain re-encode at the native frame rate and size (libx264, CRF $23$), and the third the same re-encode resampled to $8$ fps. The fourth is the full recipe, which also rescales to $512{\times}512$ and sets a GOP of $12$. Each step adds one change to the one before it, so a jump between two columns belongs to that change. ReStraV moves at the $8$ fps step and WaveRep's head at the re-encode. For WaveRep's backbone the jump comes at the last step, which adds the square resize and the GOP of $12$ and raises its rate on four collections, to as much as $65\%$, and lowers it on MSR-VTT and K400.

\begin{table}[htb]
  \centering
  \footnotesize
  \setlength{\tabcolsep}{3pt}
  \renewcommand{\arraystretch}{1.02}
  \caption{\textbf{The processing step that moves a false-positive rate depends on the detector.} Rate on real clips at a $5\%$ threshold set as shipped, mean of five seeds. Own: the training cell's real clips; others: range over the other five collections.}
  \label{tab:reprocess}
  \fitwidth{%
  \begin{tabular}{@{}llcccccccc@{}}
    \toprule
    & & \multicolumn{2}{c}{As shipped} & \multicolumn{2}{c}{H.264 re-encode} & \multicolumn{2}{c}{Plus $8$ fps} & \multicolumn{2}{c}{Recipe} \\
    \cmidrule(lr){3-4}\cmidrule(lr){5-6}\cmidrule(lr){7-8}\cmidrule(l){9-10}
    Detector & Fitted on & Own & Others & Own & Others & Own & Others & Own & Others \\
    \midrule
    ReStraV & \gvb & $.04$ & $.03$--$.28$ & $.03$ & $.04$--$.28$ & $.78$ & $.44$--$.95$ & $.78$ & $.50$--$.94$ \\
     & \aigvd & $.05$ & $.09$--$.90$ & $.05$ & $.10$--$.90$ & $.72$ & $.57$--$.72$ & $.72$ & $.56$--$.78$ \\
    \addlinespace[2pt]
    WaveRep head & \gvb & $.04$ & $.03$--$.27$ & $.05$ & $.03$--$.06$ & $.05$ & $.03$--$.08$ & $.07$ & $.04$--$.14$ \\
     & \aigvd & $.06$ & $.02$--$.24$ & $.05$ & $.02$--$.05$ & $.07$ & $.03$--$.05$ & $.12$ & $.04$--$.07$ \\
    \addlinespace[2pt]
    WaveRep backbone & \gvb & $.05$ & $.04$--$.81$ & $.04$ & $.02$--$.78$ & $.05$ & $.04$--$.75$ & $.15$ & $.06$--$.24$ \\
     & \aigvd & $.05$ & $.10$--$1.00$ & $.04$ & $.09$--$1.00$ & $.06$ & $.10$--$1.00$ & $.65$ & $.33$--$.47$ \\
    \addlinespace[2pt]
    CLIP, PCA-13 & \gvb & $.05$ & $.01$--$.17$ & $.04$ & $.00$--$.14$ & $.04$ & $.00$--$.13$ & $.10$ & $.02$--$.20$ \\
     & \aigvd & $.06$ & $.03$--$.07$ & $.05$ & $.03$--$.08$ & $.04$ & $.02$--$.08$ & $.10$ & $.05$--$.08$ \\
    \bottomrule
  \end{tabular}}
\end{table}

\para{The training cells' own pools.} The gray entries of \cref{tab:originfpr} check our re-extraction. Each detector flags $4$ to $7\%$ of its own benchmark's real clips. The one exception is itself an effect of identity. The codec-motion probe fitted on \genvideo has its threshold set on two real pools at once, and it flags almost none of MSR-VTT ($0.4\%$) but $9\%$ of Kinetics-400.

\section{Deep Detectors under the Recipe}
\label{app:recipe}

We put every clip of the \gvb and \aigvd cells through the recipe and re-extracted CLIP, WaveRep, and ReStraV with the same extractors. Of the $27{,}000$ \gvb clips, $20{,}966$ keep $K{=}12$ non-I-frames, and every MuseV and Stable Video Diffusion clip is dropped. All $2{,}284$ \aigvd clips keep $K{=}12$. To change only the processing, \cref{tab:recipe_paired} also scores the features as shipped on exactly these clips and generators. On the five surviving generators, CLIP as shipped also clears its floor, by $+0.195$. Its shortfall against the $0.15$ cut on all seven generators in \cref{tab:identity} therefore appears only when the two shortest generators, which the recipe drops, are included. ReStraV is the one feature set whose \gvb score collapses under the recipe, and the one whose \aigvd score rises.

Fitted and tested under the recipe, ReStraV and WaveRep's backbone still over-flag and WaveRep's head no longer does. \cref{tab:recipefpr} repeats the false-positive analysis of \cref{tab:originfpr} with the detectors fitted on recipe clips, thresholded at $5\%$ false positives on the recipe cell's held-out real clips and applied to every real collection through the recipe. Its rates are five-seed means, with standard deviations of at most $0.06$. Every rate of $15\%$ or more there belongs to a detector fitted on \aigvd. ReStraV flags $18$ to $43\%$ of four collections, and WaveRep's backbone flags $35\%$ of MSR-VTT and $29\%$ of Kinetics-400. Against \cref{tab:originfpr}, processing every collection alike lowers the highest rate of every deep detector. CLIP's falls from $17$ to $13\%$ and WaveRep's head's from $27$ to $9\%$. ReStraV fitted on \aigvd falls from $90$ to $43\%$, and WaveRep's backbone fitted on \aigvd from $100$ to $35\%$.

\begin{table}[htb]
  \centering
  \footnotesize
  \setlength{\tabcolsep}{4pt}
  \renewcommand{\arraystretch}{1.02}
  \caption{\textbf{The recipe changes ReStraV's score and little else on \gvb.} The same clips and generators as shipped and through the recipe, five-seed means under the shared readout.}
  \label{tab:recipe_paired}
  \fitwidth{%
  \begin{tabular}{@{}lcccccccc@{}}
    \toprule
    & \multicolumn{2}{c}{\gvb LOGO OOD} & \multicolumn{2}{c}{Floor (\rvr)} & \multicolumn{2}{c}{Margin} & \multicolumn{2}{c}{\aigvd LOGO OOD} \\
    \cmidrule(lr){2-3}\cmidrule(lr){4-5}\cmidrule(lr){6-7}\cmidrule(l){8-9}
    Detector & Shipped & Recipe & Shipped & Recipe & Shipped & Recipe & Shipped & Recipe \\
    \midrule
    CLIP, PCA-13 & $0.961$ & $0.919$ & $0.766$ & $0.752$ & $+0.195$ & $+0.168$ & $0.735$ & $0.629$ \\
    WaveRep head & $0.995$ & $0.971$ & $0.536$ & $0.539$ & $+0.458$ & $+0.432$ & $0.983$ & $0.970$ \\
    WaveRep backbone & $1.000$ & $0.984$ & $0.793$ & $0.778$ & $+0.207$ & $+0.206$ & $0.999$ & $0.987$ \\
    ReStraV & $0.852$ & $0.601$ & $0.577$ & $0.587$ & $+0.274$ & $\mathbf{+0.014}$ & $0.672$ & $0.783$ \\
    \bottomrule
  \end{tabular}}
\end{table}

\begin{table}[htb]
  \centering
  \footnotesize
  \setlength{\tabcolsep}{3pt}
  \renewcommand{\arraystretch}{1.02}
  \caption{\textbf{With every collection processed alike, ReStraV and WaveRep's backbone still over-flag.} Gray: the training cell's own pools. Bold: $0.15$ or more.}
  \label{tab:recipefpr}
  \fitwidth{%
  \begin{tabular}{@{}llcccccc@{}}
    \toprule
    Detector & Fitted on & Vript & HD-VG & \aigvd & MSR-VTT & K400 & \dvf \\
    \midrule
    CLIP & \gvb & \textcolor{gray}{$.04$} & \textcolor{gray}{$.06$} & $.13$ & $.03$ & $.02$ & $.06$ \\
    CLIP & \aigvd & $.02$ & $.01$ & \textcolor{gray}{$.06$} & $.00$ & $.00$ & $.01$ \\
    WaveRep head & \gvb & \textcolor{gray}{$.06$} & \textcolor{gray}{$.03$} & $.09$ & $.03$ & $.04$ & $.05$ \\
    WaveRep head & \aigvd & $.02$ & $.03$ & \textcolor{gray}{$.04$} & $.01$ & $.02$ & $.03$ \\
    WaveRep backbone & \gvb & \textcolor{gray}{$.03$} & \textcolor{gray}{$.04$} & $.09$ & $.12$ & $.10$ & $.06$ \\
    WaveRep backbone & \aigvd & $.10$ & $.07$ & \textcolor{gray}{$.03$} & $\mathbf{.35}$ & $\mathbf{.29}$ & $.09$ \\
    ReStraV & \gvb & \textcolor{gray}{$.05$} & \textcolor{gray}{$.07$} & $.13$ & $.01$ & $.04$ & $.09$ \\
    ReStraV & \aigvd & $\mathbf{.18}$ & $\mathbf{.20}$ & \textcolor{gray}{$.06$} & $\mathbf{.43}$ & $.12$ & $\mathbf{.19}$ \\
    \bottomrule
  \end{tabular}}
\end{table}

\cref{tab:sides} splits each cross-benchmark loss of \cref{sec:sides} into its two sides. Its left half uses the cells' own clips, as shipped for the deep detectors and under the recipe for the codec-motion probe, so its rates differ from \cref{tab:originfpr} by up to $0.04$. The home true-positive rate is measured on the training cell, and the false-positive rate, true-positive rate, and AUC on the target. A target false-positive rate above $0.05$ marks a loss on the real side, and a target true-positive rate below the home rate marks a loss on the generated side. The codec-motion probe and CLIP fitted on \gvb lose on both sides. ReStraV fitted on \gvb loses on the generated side alone, and WaveRep's head stays within a few points on both.

The right half of \cref{tab:sides} repeats this split with every clip of both cells through the recipe, as five-seed means. ReStraV fitted on \aigvd then loses on both sides, flagging $20\%$ of \gvb's real clips and catching $37\%$ of its generated ones against $50\%$ at home. ReStraV fitted on \gvb changes side. Its home rate falls to $43\%$ under the recipe, so it no longer loses on \aigvd's generated clips ($42\%$), and it flags $12\%$ of \aigvd's real clips instead. WaveRep's backbone flags $11$ to $12\%$ of the target's real clips in both directions. WaveRep's head stays closest to its home rates under both forms of the clips.

\section{Removing Identity with Real Videos of the Target}
\label{app:realonly}

Removing identity directions with real videos of the target fails its success rule, while recalibrating the threshold already brings the false-positive rate back to about $5\%$. Half of the target's real clips is the only target data any method sees, and the other half and every generated clip form the test set. The method projects out the directions that separate the training benchmark's real clips from that half. It repeats until held-out real clips of the two are no longer separable (AUC below $0.55$). It then refits the detector on the projected training features and sets its threshold on the target's real clips. We compare it with recalibrating the threshold alone and with correlation alignment (CORAL), which aligns the covariance of the training features to that of the target's real clips. The success rule was fixed before the run: beat the best baseline by $0.02$ mean AUC, and beat no adaptation in at least $70\%$ of cases. Every adaptation brings the false-positive rate back to about $5\%$, which recalibrating the threshold already does. On ranking (\cref{tab:realonly}), identity removal helps the codec-motion probe, loses to CORAL on CLIP and ReStraV, and ties at ceiling on both WaveRep variants, so it fails the rule. Removing only the directions that separate \gvb's own two real pools drives that separability to chance (CLIP $0.85$ to $0.55$). It leaves the distance to \aigvd's real pool almost unchanged ($0.93$ to $0.92$), so identity in these features is specific to each pair of collections.

\begin{table}[htb]
  \centering
  \footnotesize
  \setlength{\tabcolsep}{3.5pt}
  \renewcommand{\arraystretch}{1.02}
  \caption{\textbf{A transfer loss can fall on either class.} Each detector at its home $5\%$ threshold, applied to the other benchmark, with the clips as shipped (left) and through the recipe (right). GVB: \gvb; AIGVD: \aigvd. Bold: FPR of $0.15$ or more.}
  \label{tab:sides}
  \fitwidth{%
  \begin{tabular}{@{}llcccccccc@{}}
    \toprule
    & & \multicolumn{4}{c}{As shipped} & \multicolumn{4}{c}{Every clip through the recipe} \\
    \cmidrule(lr){3-6}\cmidrule(l){7-10}
    Detector & Fitted $\to$ applied & Home TPR & FPR & TPR & AUC & Home TPR & FPR & TPR & AUC \\
    \midrule
    Codec-motion & GVB$\to$AIGVD & $.29$ & $\mathbf{.17}$ & $.14$ & $.49$ & -- & -- & -- & -- \\
     & AIGVD$\to$GVB & $.10$ & $\mathbf{.28}$ & $.22$ & $.46$ & -- & -- & -- & -- \\
    CLIP, PCA-13 & GVB$\to$AIGVD & $.67$ & $\mathbf{.15}$ & $.42$ & $.66$ & $.74$ & $\mathbf{.16}$ & $.33$ & $.62$ \\
     & AIGVD$\to$GVB & $.32$ & $.04$ & $.38$ & $.79$ & $.22$ & $.02$ & $.08$ & $.73$ \\
    ReStraV & GVB$\to$AIGVD & $.92$ & $.05$ & $.22$ & $.57$ & $.43$ & $.12$ & $.42$ & $.67$ \\
     & AIGVD$\to$GVB & $.59$ & $.13$ & $.56$ & $.81$ & $.50$ & $\mathbf{.20}$ & $.37$ & $.67$ \\
    WaveRep head & GVB$\to$AIGVD & $.97$ & $.06$ & $.93$ & $.99$ & $.89$ & $.09$ & $.89$ & $.97$ \\
     & AIGVD$\to$GVB & $.92$ & $.04$ & $.97$ & $1.00$ & $.82$ & $.03$ & $.83$ & $.97$ \\
    WaveRep backbone & GVB$\to$AIGVD & $1.00$ & $.05$ & $.99$ & $1.00$ & $1.00$ & $.12$ & $.91$ & $.96$ \\
     & AIGVD$\to$GVB & $1.00$ & $.11$ & $1.00$ & $1.00$ & $.97$ & $.11$ & $.93$ & $.97$ \\
    \bottomrule
  \end{tabular}}
\end{table}

\cref{tab:recal} gives the rates behind the recalibration of \cref{sec:realonly}. The threshold is set at $5\%$ false positives on half of the target's real clips. The codec-motion probe covers all twelve ordered pairs over four cells, and the deep detectors the two directions between \gvb and \aigvd. The seed-mean false-positive rate lies between $4.1$ and $6.8\%$ for every detector and pair of cells, and single runs vary from $2.1$ to $12.1\%$, most where the half holds only $231$ clips. Recalibration costs the codec-motion probe and ReStraV the most true positives.

\section{Per-Detector Readout Details}
\label{app:harness}

We re-fit every detector's features under the shared readout, so that a difference between two scores is a difference between representations. This section gives the setup behind the rows of \cref{tab:identity}.

\para{WaveRep.} WaveRep~\citep{corvi2025waverep} is a DINOv2 ViT-B/14 backbone with a one-output linear head, released as one checkpoint trained on four generators, which its paper lists as Pyramid-Flow, CogVideoX-1.5, Allegro, and Open-Sora-Plan. We follow its demo code for the per-frame transform and score $16$ frames sampled uniformly from a $2$\,s window at the clip midpoint, the window ReStraV uses. The head's per-frame output is a log-likelihood ratio, and WaveRep's head in our tables is the shared readout on its mean, median, and maximum over the $16$ frames. WaveRep's backbone is the shared readout on the $768$-dimensional class-token feature before the head, averaged over the same frames. The checkpoint is never re-trained.

\para{FVMD.} FVMD's full $1024$-dimensional descriptor (concatenated velocity and acceleration histograms from a PIPs++ point tracker~\citep{zheng2023pointodyssey}) is read through the shared readout, and a $13$-dimensional PCA variant is in \cref{tab:supp_seeds_baselines}.

\begin{table}[htb]
  \centering
  \footnotesize
  \setlength{\tabcolsep}{4pt}
  \renewcommand{\arraystretch}{1.02}
  \caption{\textbf{Adapting with unlabeled real videos of the target.} Mean AUC on the target cell over every ordered pair of training and target cells (codec-motion probe: twelve pairs over four cells; deep detectors: \gvb$\leftrightarrow$\aigvd), five seeds.}
  \label{tab:realonly}
  \fitwidth{%
  \begin{tabular}{@{}lccc@{}}
    \toprule
    Detector & No adaptation & CORAL & Identity removal \\
    \midrule
    Codec-motion probe & $0.596$ & $0.586$ & $\mathbf{0.641}$ \\
    CLIP, PCA-13 & $0.725$ & $\mathbf{0.740}$ & $0.716$ \\
    ReStraV & $0.694$ & $\mathbf{0.703}$ & $0.641$ \\
    WaveRep head & $0.991$ & $0.991$ & $0.991$ \\
    WaveRep backbone & $0.998$ & $\mathbf{0.999}$ & $0.998$ \\
    \midrule
    Mean & $0.801$ & $\mathbf{0.804}$ & $0.797$ \\
    \bottomrule
  \end{tabular}}
\end{table}

\para{RAFT.} The RAFT row decodes $12$ frames at $256{\times}256$ and runs pretrained RAFT-Large~\citep{teed2020raft} on the $11$ consecutive frame pairs. The model has $5.26$M parameters, and we use the torchvision weights trained on FlyingChairs, FlyingThings3D, Sintel, KITTI, and HD1K. The dense flow is average-pooled onto the same $32{\times}32$ grid that the codec motion vectors occupy. The pooled $(\Delta x, \Delta y)$ field is fed to the exact $13$-statistic pipeline of \cref{app:probes}, so any difference between the codec-motion and RAFT rows reflects the motion source alone, codec block matching against learned dense flow.

\para{ReStraV.} ReStraV~\citep{restrav2024} summarizes each clip's per-frame DINOv2~\citep{oquab2024dinov2} ViT-S/14 embedding sequence as a $21$-dimensional trajectory descriptor (early stepwise distances and turning angles between consecutive embedding deltas, $T{=}24$ frames from a $2$\,s window at the clip midpoint). At $3$ fps, as on MSR-VTT, the window holds about seven distinct frames, so most of the $24$ are repeats. ReStraV's own head is a small multi-layer perceptron. Under the shared readout the $21$ features reach LOGO OOD $0.931$ at a floor of $0.586$. Across six readouts on the same features, the OOD ranges over $0.882$ to $0.931$ and the floor over $0.586$ to $0.635$, so the readout moves this detector's floor by about five points.

\para{DeMamba.} DeMamba~\citep{chen2024demamba} releases training code but no checkpoints and no inference path, so it is surveyed only.

\para{D3.} D3~\citep{ma2024d3} takes per-frame X-CLIP ViT-B/16~\citep{ni2022xclip} embeddings and the second-order temporal difference $\Delta^2 f(t) = f(t{+}1) - 2f(t) + f(t{-}1)$, and its features averaged over time give the $768$-dimensional row of \cref{tab:identity}.

\end{document}